\documentclass{article}
\usepackage{iclr2027_conference,times}

\usepackage{amsmath,amsfonts,bm}

\usepackage{amssymb}

\usepackage{hyperref}
\usepackage{url}
\usepackage{graphicx}
\usepackage{booktabs}
\usepackage{enumitem}
\usepackage{wrapfig}
\usepackage{needspace}

\newcommand{\methodname}{\textsc{Sherpa}}
\newcommand{\resulttablesetup}{\small\setlength{\tabcolsep}{4.5pt}}
\newcommand{\prefNone}{Non.}
\newcommand{\prefAtt}{Att.}
\newcommand{\prefSub}{Sub.}
\newcommand{\prefCtr}{Ctr.}
\newcommand{\prefCau}{Cau.}
\newcommand{\prefStp}{Stp.}
\newcommand{\prefVer}{Ver.}
\newcommand{\prefOvl}{Ovl.}

\title{\methodname: Teaching LLMs to Teach Adaptively}
\iclrfinalcopy
\author{Weixian Xu\thanks{Project done while Weixian Xu is visiting Stanford.}\\
Stanford University \\
\texttt{weixianx@stanford.edu}
\And
Yanzhe Zhang\\
Georgia Tech \\
\texttt{z\_yanzhe@gatech.edu}
\And
Zora Zhiruo Wang\\
Carnegie Mellon University \\
\texttt{zhiruow@cs.cmu.edu}
\AND
Changyu Chen\\
Stanford University\\
\texttt{chency@stanford.edu}
\And
Diyi Yang\\
Stanford University \\
\texttt{diyiy@stanford.edu}
}

\begin{document}
\maketitle

\begin{abstract}
Large language models (LLMs) have become increasingly capable problem solvers, but being able to solve a problem is not the same as being able to teach it. Existing approaches to training LLMs as teachers rely on demonstrations, preference data, or predefined pedagogical criteria that specify what good teaching looks like. However, these signals are often not grounded in individual student learning outcomes, where effective teaching strategies can vary substantially across learners.
To address this, we introduce \methodname{}\footnote{Named after the Sherpa people, renowned Himalayan mountaineering guides, our framework aims to enable language models to help people reach new heights through learning.}, a multi-turn reinforcement learning framework that instantiates multiple student archetypes with LLMs conditioned on distinct learning preferences and trains a teacher model to adapt its instruction by directly maximizing their learning outcomes.
Teacher LLMs trained with \methodname{} improve instructed students' performance across all archetypes by an average of 20.5 percentage points.
Under MathTutorBench's evaluation, \methodname{} raises the overall pedagogy score from 52.5\% to 79.2\%, indicating better teaching responses.
Our human studies show that the trained teacher is preferred over the base model in 79.6\% of pairwise comparisons.
Together, \methodname{} trains LLM teachers to adapt to diverse simulated students and become better aligned with human teachers, paving the road towards AI tutors teaching real students.
\footnote{Code and model are available at \url{https://github.com/SALT-NLP/Sherpa}.}
\end{abstract}

\section{Introduction}
\label{sec:introduction}

Large language models (LLMs) have advanced rapidly, achieving strong capabilities in mathematical reasoning and code generation \citep{yang2024qwen25math,hui2024qwen25coder}. Solving a problem, however, is different from teaching someone else to solve it. \citet{srinivasa2025tutorbench} finds that none of the 16 evaluated frontier models scores above 56\% against its expert-written tutoring rubrics, while \citet{macina-etal-2025-mathtutorbench} shows that strong problem-solving ability does not directly translate into effective teaching. This gap matters for human learners: studies of general-purpose AI use have raised concerns that receiving effective task assistance does not necessarily translate into learning, and may even reduce cognitive engagement or subsequent independent performance \citep{kosmyna2025brain,bastani2025guardrails}. LLMs may thus erode learners' agency rather than help preserve it. Developing models that help people learn to think and solve problems independently is therefore increasingly important, beyond improving individual dimensions of teaching skill.

One approach to AI-assisted learning is to build educational systems around existing LLMs, injecting human-designed pedagogical strategies through prompts or external scaffolding \citep{chen2024chattutor,puech2024steering}. Such systems can provide useful instructional structure, but they primarily shape the model's behavior at inference time. To \emph{develop teaching as an intrinsic capability of the model itself}, a growing body of work instead trains LLMs specifically as teachers. Supervised fine-tuning and preference optimization came first, but they rely on the quality and coverage of instructional demonstrations or preference data \citep{macina-etal-2023-mathdial,scarlatos2025training}. More recent work has turned to reinforcement learning so that teachers can learn from their own interactions with simulated students \citep{dinucu-jianu-etal-2025-problem,chang2026pearl}. Yet these approaches share a fundamental limitation: their training signals encode predefined notions of what good teaching looks like, rather than being grounded directly in diverse students' learning outcomes. In practice, students differ in their preferences and needs, and effective teaching cannot be fully specified without accounting for this heterogeneity.

To close this gap, we introduce \methodname{}, a multi-turn reinforcement learning framework that trains a teacher to improve student task-solving across diverse student archetypes. In this work, we train through interactions with language-model students and use pairwise comparisons from human teachers to assess whether the learned teaching behavior aligns with human judgments of effective teaching. Rather than predefining what constitutes good teaching, \methodname{} separates pedagogical specification from teacher optimization: diverse instructional needs are instantiated on the student side, while the teacher is optimized based on student improvement. Each teaching episode follows a preparation–tutoring–testing cycle: given a problem, the teacher first prepares a solution, tutors the student over multiple turns of interaction, and is rewarded based on the student's improvement at test time. However, student models exhibit similar preferences for instructional strategies even after prompting or across model families, making them insufficiently diverse as archetypes for our environment. We therefore design a gating mechanism that makes instructional strategies consequential: guidance that meets the student's preference passes the gate and is forwarded to the student backbone model to elicit substantive reasoning, whereas incompatible guidance is blocked and receives feedback requesting more appropriate instruction. The teacher must therefore infer each student's needs from interaction and adapt its teaching accordingly. In this way, \methodname{} allows diverse teaching behaviors to emerge while keeping the optimization objective simple and outcome-based.

We identify six important student learning preferences from education literature and initiate corresponding student archetypes using Qwen3-1.7B \citep{yang2025qwen3technicalreport} as the backbone model, and train a Qwen3-8B \citep{yang2025qwen3technicalreport} teacher to help students on MATH problems \citep{hendrycksmath2021}.
\methodname{} improves final test accuracy across all student archetypes, including those unseen during training, increasing average student performance by 20.5 percentage points. Moreover, these gains transfer to unseen data and settings: on MathTutorBench~\citep{macina-etal-2025-mathtutorbench}, our training raises the overall pedagogy score from 52.5\% to 79.2\%, outperforming PedagogicalRL \citep{dinucu-jianu-etal-2025-problem} on all four pedagogical metrics and achieving the highest average pedagogy score among the evaluated models. Our human study shows that the trained teacher is preferred over the base model in 79.6\% of pairwise comparisons.
Finally, we compare different student archetype designs and show that our design effectively elicits the student behaviors needed for our training environment compared to baselines. We further ablate different reward designs and show that a simple optimization objective substantially improves generalization.

\vspace{-0.01in}
\section{How to Train an Adaptive Teacher LLM?}
\label{sec:method}
\vspace{-0.01in}

\begin{figure}[t]
\centering
\includegraphics[width=0.9\linewidth]{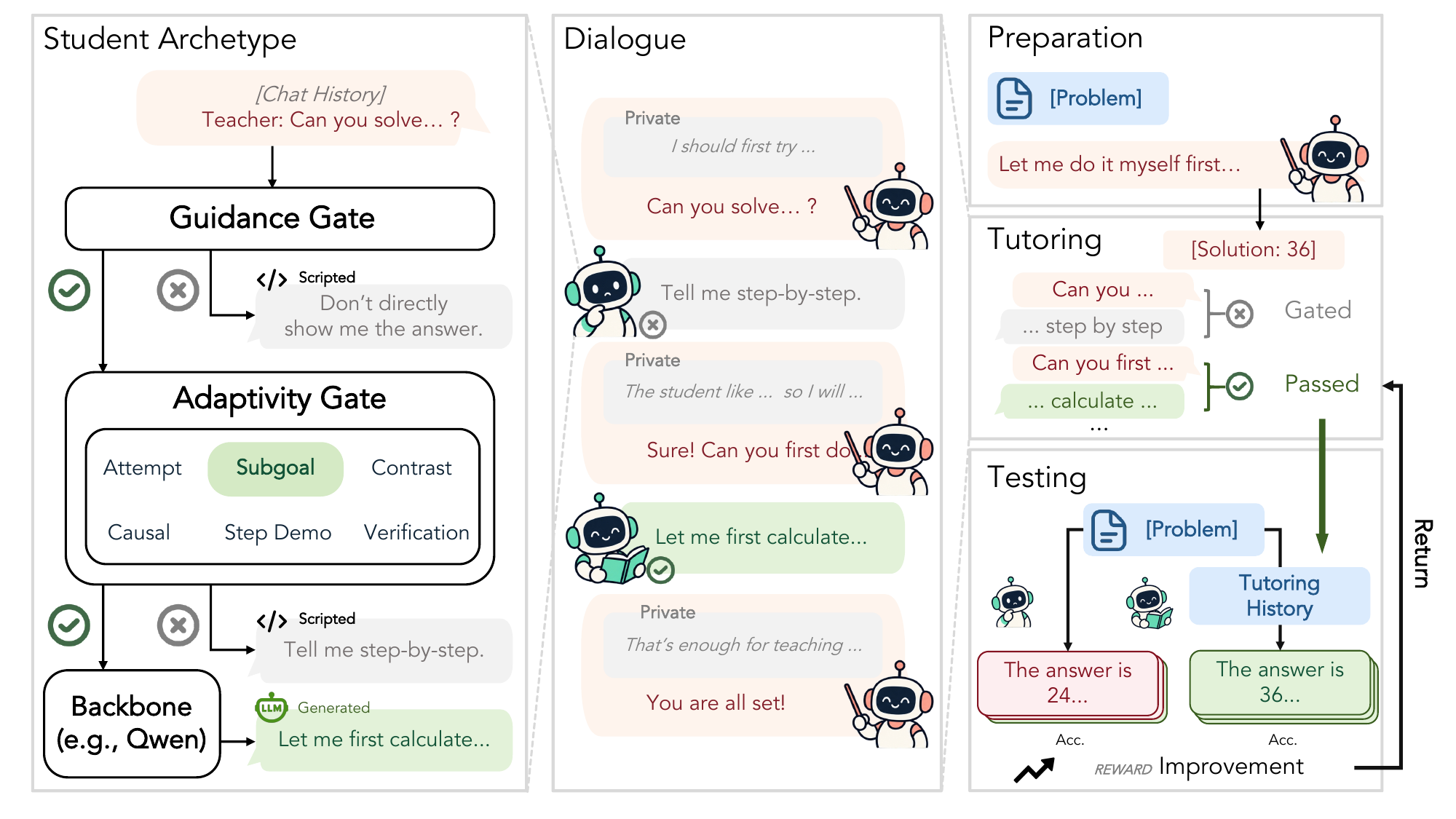}
\caption{Overview of our framework. The teacher interacts with student archetypes through preparation, tutoring, and testing, and receives a reward for tutoring based on the student's improvement. The student archetype is implemented by a backbone model with a gating mechanism.}
\label{fig:pipeline}
\end{figure}
\vspace{-0.01in}

As shown in Figure~\ref{fig:pipeline}, our framework trains an adaptive teacher LLM through interaction with diverse student archetypes. The teaching process follows a classroom cycle with three stages, which together constitute a teaching episode: \textbf{preparation}, in which the teacher prepares the material; \textbf{tutoring}, in which the teacher chats with the student; and \textbf{testing}, in which the student completes a test.
Based on this design, we optimize the teacher via multi-turn reinforcement learning. We first formulate the problem (\S\ref{sec:problem}), then describe our student archetype design (\S\ref{sec:environment}), and finally present our teacher training method (\S\ref{sec:training}).

\subsection{Problem Formulation}
\label{sec:problem}

Our goal is to train a teacher policy $\pi_\theta$ to teach multiple student archetypes, each student $\mathcal{E}_z$ defined by a latent student type $z$ that determines its instructional demands.
Here, we use archetypes to capture representative instructional preferences, viewing the more nuanced preferences encountered in practice as combinations of these archetypal patterns.
The teacher does not observe $z$ directly, and infers the student's needs from the observed replies.
Given a problem $x$ and student archetype $\mathcal{E}_z$, the teacher first produces a solution draft during preparation.
When entering tutoring, let $C_t^{T}$ and $C_t^{S}$ denote the teacher and student contexts at turn $t$, respectively. The initial teacher context $C_0^{T}$ contains the teacher system prompt (which includes problem $x$), the preparation request, the solution draft, and the prompt to begin tutoring. The initial student context $C_0^{S}$ contains only the student system prompt $\mathrm{sys}^{S}$.
At each turn, the teacher generates an instruction $a_t$, and the student archetype returns a reply $o_t$. We describe how $C_t^{T}$ and $C_t^{S}$ are updated in Sec~\ref{sec:environment} and Sec~\ref{sec:rollout}.

The objective is to maximize the student's improvement on $x$.
Let $p(x,z,C_H^{S})$ denote the probability that student archetype $\mathcal{E}_z$ answers $x$ correctly with its context after $H$ turns, and $p(x,z,C_0^{S})$ the probability for the same frozen student model with no interaction history. We optimize
\begin{equation}
\max_\theta\;
\mathbb E_{(x,z)\sim\mathcal D,\,
C_H^{S}\sim P_\theta(\cdot\mid x,z)}
\left[
p(x,z,C_H^{S})
-
p(x,z,C_0^{S})
\right],
\label{eq:teaching-objective}
\end{equation}
where $\mathcal D$ is the distribution over problems and student archetypes, and $P_\theta$ is the trajectory distribution induced by the teacher policy and student archetype.\footnote{Because the student backbone is frozen, $p(x,z,C_0^{S})$ is independent of $\theta$, so maximizing expected improvement is equivalent to maximizing expected post-tutoring success.}

\subsection{Student Archetype}
\label{sec:environment}

To design student archetypes with different latent variables $z$ that genuinely require different guidance strategies, we propose two principles:

\begin{enumerate}[leftmargin=*]
    \item \textbf{Selective:} student improvement should depend on whether the teacher's instruction meets the demands represented by $z$.
    \item \textbf{Informative:} student replies should reflect the preference represented by $z$.
\end{enumerate}

Together, these principles make the student's latent demands both consequential for learning and inferable from interaction. Satisfying both, however, is challenging. Existing approaches often start from models capable of solving the target problems and use prompting or unlearning to induce different student behaviors \citep{scarlatos2026simulated,song2026novice}. Such students may still benefit from generic explanations regardless of their assigned needs, while knowledge targeted by unlearning can remain accessible \citep{lynch2024eight,thaker2025unlearning}. Directly training models to mimic student behaviors can instead lead to excessive error imitation and degraded problem-solving ability \citep{sonkar2024paradox}, complicating the measurement of teaching effectiveness.

In \methodname{}, we instead use a weak model that cannot solve the problem unaided as the student backbone. This makes learning gains easier to attribute to tutoring, but the weaker model is also less reliable at expressing distinct behaviors through prompting alone. We therefore use gates to control which teacher instructions reach the student backbone. Specifically, we decompose adaptive teaching into two components: providing useful guidance and matching that guidance to the student's latent demands, and implement it with a \textbf{guidance gate} and an \textbf{adaptive gate}.

\noindent
\textbf{Guidance Gate.}
We view guidance as helping the student make progress while leaving room for the student to think. A teacher model can instead solve the problem entirely, allowing the student to reproduce it without reasoning. To prevent this shortcut, we require the teacher to guide the student toward the solution without disclosing the final answer. Prior work uses LLM judges to detect answer leakage or assess teaching quality \citep{dinucu-jianu-etal-2025-problem}, but broad judgments may either miss localized disclosure or penalize legitimate guidance such as intermediate calculations and worked steps. We therefore make the criterion explicit: the \textbf{guidance gate} uses an LLM checker to determine whether each teacher instruction reveals the ground-truth final answer. Instructions that do so are blocked; intermediate reasoning and partial solutions remain allowed. This design permits substantive guidance while leaving the student to work out the final answer.

\noindent
\textbf{Adaptive Gate.}
Avoiding answer disclosure does not by itself guarantee adaptive teaching. If every student archetype benefits from the same explanation, the teacher has no incentive to infer individual needs. We therefore assign students different preferences for teaching strategies drawn from widely studied pedagogical interventions:

\begin{itemize}[leftmargin=*]
    \item \textbf{Attempt diagnosis:} identify and explain a specific issue in the student's latest attempt \citep{metcalfe2017learning,daheim-etal-2024-stepwise}.
    \item \textbf{Causal justification:} explain why a mathematical step or claim holds \citep{chi2008observing}.
    \item \textbf{Subgoal decomposition:} identify the current subgoal and relate it to the overall solution \citep{margulieux2016employing}.
    \item \textbf{Step demonstration:} demonstrate and explain one new step before returning control to the student \citep{barbieri2023worked}.
    \item \textbf{Independent verification:} check a claim through an alternative route \citep{polya1945solve}.
    \item \textbf{Contrastive comparison:} compare plausible alternatives and explain their decisive difference \citep{schwartz1998time,alfieri2013learning}.
\end{itemize}

Unlike the guidance demands shared by all student archetypes, these different demands for teaching strategies distinguish their latent variables $z$. The teacher does not know the student's preferences before the interaction and therefore needs to adapt their teaching approach based on the feedback. We provide the full prompts used to implement the different adaptive gates in Appendix~\ref{sec:adaptive-gate-prompts}.

\noindent
\textbf{Formal Gating Mechanism.} 
At turn $t$, a teacher instruction $a_t$ is accepted only if it passes both the guidance and adaptive gates. Let $m_t\in\{0,1\}$ be the gate-acceptance indicator, with $m_t=1$ when $a_t$ passes both gates and $m_t=0$ when it is rejected. 
When $m_t=0$, the student archetype directly returns a scripted reply $o_t$ stating why the instruction failed (e.g., ``\textit{Don't directly show me the answer.}'', ``\textit{Tell me the step-by-step plan.}'').
When $m_t=1$, the student backbone generates $o_t$ based on $C_t^{S}$ and $a_t$.
The update rule for the student context $C^S_{t}$ is:
\begin{equation}
C_0^{S}=\mathrm{sys}^{S},\qquad
C_{t+1}^{S}=
\begin{cases}
C_t^{S}, & m_t=0,\\
C_t^{S}\oplus(a_t,o_t), & m_t=1.
\end{cases}
\label{eq:student-context}
\end{equation}
where $\oplus$ denotes the appending of an exchange. Only accepted exchanges reach the student. The scripted reason stays in the teacher context so that the policy can condition on it at the next turn.

\subsection{Training the Teacher Policy}
\label{sec:training}

We first describe how we roll out the teaching episode from the teacher's perspective and how we optimize the teacher based on student improvement.

\noindent
\textbf{Teaching Episode}
\label{sec:rollout}
\textit{(i) Preparation.} The teacher privately solves the assigned problem and retains a solution draft as background knowledge, just like human teachers prepare a problem's solution and underlying reasoning before teaching it. For LLM teachers, particularly small models, keeping the solution draft available throughout tutoring helps reduce drift as the dialogue grows and improves teaching performance. We provide the detailed comparison in Appendix~\ref{sec:preparation-evaluation}.
\textit{(ii) Tutoring.} The teacher initiates the dialogue with a student archetype, with the problem and the solution draft at the beginning of the teacher context.
At each turn, the teacher generates a complete response $y_t \sim \pi_\theta(\cdot\mid C_t^{T})$, consisting of private reasoning and a student-visible instruction $a_t$.
The private part allows the teacher to reason freely without being constrained by student preferences.
After receiving student reply $o_t$, the teacher context is always updated: $C_{t+1}^{T}=C_t^{T}\oplus(a_t,o_t)$.
We set a turn budget for tutoring, while the teacher may also end the tutoring earlier, mirroring the conclusion of a lesson.
\textit{(iii) Testing.} After tutoring, we test the student by providing $C_H^{S}$ together with the original problem, and asking it to produce a complete solution. We estimate accuracy through repeated testing.
To measure the student's unaided performance, we evaluate the same student model without providing the interaction history accumulated during tutoring.

\noindent
\textbf{Reward}
\label{sec:reward-design}
For a tutoring trajectory on problem $x$ with student archetype $\mathcal{E}_z$, the trajectory reward $R$ is the difference between the empirical estimators of $p(x,z,C_H^{S})$ and $p(x,z,C_0^{S})$. We obtain these estimates by repeatedly testing the student with and without interaction history, respectively.

\noindent
\textbf{Algorithm}
\label{sec:algorithm}
Our algorithm design builds on critic-free RL algorithms \citep{shao2024deepseekmath, liu2025gem}. 
For each problem, we sample a group of $G$ trajectories sharing the student archetype and preparation context, each with a reward $R_i$.
Since rejected turns do not enter $C_H^{S}$, they make no contribution to student improvement (Sec \ref{sec:environment}).
We therefore apply a mask to their turn-level returns without making any assumption about their quality.
Let $m_{i,t}\in\{0,1\}$ indicate whether turn $t$ in trajectory $i$ passes the guidance and adaptive gates. We define the masked return as $U_{i,t}=m_{i,t}R_i$.
This design admits a local surrogate-objective interpretation: under the rollout policy, $U_{i,t}$ serves as the return weight in a masked policy-gradient surrogate. We provide a description of the corresponding objective and its derivation in Appendix~\ref{app:masked-objective}, and separately show the effects of student-improvement rewards and turn-level masking in Appendix~\ref{sec:mask-ablation}.

Because the return varies across turns and trajectories, we compare returns at the same turn within each group. Let $\overline U_{-i,t}$ denote the mean return of the other trajectories reaching turn $t$. Leave-one-out centering \citep{ahmadian2024back} and turn-level normalization yield

\begin{equation}
\widetilde A_{i,t}
=
U_{i,t}-\overline U_{-i,t},
\qquad
A_{i,t}
=
\frac{\widetilde A_{i,t}}
{\sigma_{\mathcal B}+\varepsilon}
+\ell_{i,t},
\label{eq:turn-advantage}
\end{equation}

where $\sigma_{\mathcal B}$ is the batch standard deviation of the centered returns, $\varepsilon$ is a numerical stabilizer, and $\ell_{i,t}$ contains the local format and repetition penalties, added after normalization to preserve their scale and improve training stability.
When a trajectory is the only one reaching turn $t$, no leave-one-out comparison is available; we therefore set $\widetilde A_{i,t}=0$, leaving only the local penalties.
We broadcast $A_{i,t}$ to all tokens in the teacher response $y_{i,t}$, including private reasoning and the student-visible instruction, and optimize the teacher using asynchronous PPO, with further details of the implementation provided in Appendix~\ref{app:async-implementation}.

\section{Adapting Teaching to Individuals}
\label{sec:results}

\subsection{Teaching Diverse Archetypes}
\label{sec:experimental-setup}

\noindent
\textbf{Student Archetypes.}
We instantiate seven student archetypes: six with distinct preferences enforced by their respective adaptive gates, and one without preference (None, \prefNone). The descriptions and names below distinguish them by their preferences. Our model is trained on four of these archetypes (None, \prefNone; Attempt, \prefAtt; Subgoal, \prefSub; Contrast, \prefCtr), making them in-distribution (ID) and the remaining three (Causal, \prefCau; Step, \prefStp; and Verification, \prefVer) for out-of-distribution (OOD) evaluation. Detailed descriptions of the student preferences are provided in Section~\ref{sec:environment}.

\noindent
\textbf{Problems Filtering.}
We focus on mathematics, where teacher models possess substantial task knowledge and student answers can be easily judged. Our problems come from MATH \citep{hendrycksmath2021}, with filtering to remove less teachable cases: problems beyond the teacher's capabilities and those the student can already solve unaided. Therefore, we select problems for which the teacher has relevant knowledge and the student has room to improve. This filtering retains 759 training and 528 evaluation problems from the corresponding dataset splits.

\noindent
\textbf{Settings.}
We use Qwen3-8B \citep{yang2025qwen3technicalreport} as the teacher backbone and Qwen3-1.7B as the student backbone.
For evaluation, we compare with the untrained Qwen3-8B backbone and a Qwen3-8B trained with PedagogicalRL.
We also compare \methodname{}\textsubscript{DP}, trained with diverse-preference students (all ID archetypes), and \methodname{}\textsubscript{NP}, trained with no-preference students (None).
See Appendix~\ref{sec:results-evaluation-details} for training and evaluation details.

\begin{table}[!t]
\centering\resulttablesetup
\caption{Student test accuracy after teaching (\%; higher is better). The student's overall unaided accuracy is 36\%. \methodname{}\textsubscript{NP} and \methodname{}\textsubscript{DP} are trained with no-preference and diverse-preference students, respectively. \prefNone, \prefAtt, \prefSub, and \prefCtr\ are in-distribution (ID); \prefCau, \prefStp, and \prefVer\ are out-of-distribution (OOD). Overall (\prefOvl) is the average across all seven archetypes. \textbf{Bold} values indicate the best result in each column. The final column reports 95\% confidence intervals for overall accuracy.
}
\label{tab:retest-accuracy}
\begin{tabular}{lccccccccccc}
\toprule
& \multicolumn{4}{c}{ID} & \multicolumn{3}{c}{OOD} & \multicolumn{3}{c}{Average} & \\
\cmidrule(lr){2-5}\cmidrule(lr){6-8}\cmidrule(lr){9-11}
\multicolumn{1}{l}{Teacher} & \multicolumn{1}{c}{\prefNone} & \multicolumn{1}{c}{\prefAtt} & \multicolumn{1}{c}{\prefSub} & \multicolumn{1}{c}{\prefCtr} & \multicolumn{1}{c}{\prefCau} & \multicolumn{1}{c}{\prefStp} & \multicolumn{1}{c}{\prefVer} & \multicolumn{1}{c}{ID} & \multicolumn{1}{c}{OOD} & \multicolumn{1}{c}{\prefOvl} & \multicolumn{1}{c}{95\% CI} \\
\midrule
Qwen3-8B & 53.2 & 50.7 & 48.4 & 44.1 & 54.4 & 47.8 & 41.5 & 49.1 & 47.9 & 48.6 & [46.1, 51.0] \\
PedagogicalRL & 63.0 & 59.4 & 54.7 & 44.4 & 59.0 & 44.3 & 43.8 & 55.4 & 49.1 & 52.7 & [50.3, 55.1] \\
\midrule
\methodname{}\textsubscript{NP}& 63.1 & 58.8 & 52.0 & 45.3 & 55.1 & 40.6 & 49.2 & 54.8 & 48.3 & 52.0 & [49.7, 54.5] \\
\methodname{}\textsubscript{DP} & \textbf{76.2} & \textbf{78.2} & \textbf{72.7} & \textbf{60.7} & \textbf{77.1} & \textbf{51.8} & \textbf{66.4} & \textbf{72.0} & \textbf{65.1} & \textbf{69.0} & [66.8, 71.2] \\
\bottomrule
\end{tabular}
\end{table}

\noindent
\textbf{Results.}
As shown in Table~\ref{tab:retest-accuracy}, \methodname{}\textsubscript{DP} achieves an overall test accuracy of 69.0\%, outperforming the untrained backbone, PedagogicalRL, and \methodname{}\textsubscript{NP} on every student archetype. \methodname{}\textsubscript{NP} also improves on the untrained backbone, raising overall test accuracy from 48.6\% to 52.0\%, comparable to PedagogicalRL's 52.7\%. This demonstrates the effectiveness of using student improvement to guide teacher training. Furthermore, \methodname{}\textsubscript{DP} performs substantially better than \methodname{}\textsubscript{NP} across student archetypes, demonstrating the importance of diverse student archetypes in training. The gains in accuracy differ across archetypes, indicating that improvements do not transfer equally across them (Causal: 55.1\% to 77.1\%, Step: 40.6\% to 51.8\%). Additional results on different student backbone models and multiple training seeds are provided in Appendices~\ref{sec:student-backbone-stability} and~\ref{sec:multiseed}.

\subsection{Evaluation Beyond the Student Archetype}

\noindent
\textbf{Setup}
To test whether our gains generalize beyond our setting, we evaluate on MathTutorBench~\citep{macina-etal-2025-mathtutorbench}, a teacher-grounded mathematics tutoring benchmark.
It covers math expertise (Problem Solving, PS; Socratic Questioning, SQ), student understanding (Solution Correctness, SC; Mistake Location, ML; Mistake Correction, MC), and teacher response generation (Scaffolding, Scaf.; Instruction Following, IF; and harder variants Scaf.-H and IF-H). Because the last category most directly measures teaching quality, we summarize it with Pedagogy Average (Ped.\ Avg.), the mean of its four tasks.
Evaluation details are in Appendix~\ref{sec:mathtutorbench-evaluation-details}.

\begin{table}[t]
\centering\resulttablesetup
\caption{MathTutorBench scores (\%; higher is better). Ped.\ Avg.\ is the average score across teacher response generation tasks. \textbf{Bold} values indicate the best result in each column.
}
\label{tab:mathtutorbench}
\begin{tabular}{lcccccccccc}
\toprule
& \multicolumn{2}{c}{Math Expertise} & \multicolumn{3}{c}{Student Understanding} & \multicolumn{4}{c}{Teacher Response Generation} & \\
\cmidrule(lr){2-3}\cmidrule(lr){4-6}\cmidrule(lr){7-10}
Teacher & PS & SQ & SC & ML & MC & Scaf. & IF & Scaf.-H & IF-H & Ped.\ Avg. \\
\midrule
Qwen3-8B & 92.1 & \textbf{25.3} & 70.6 & \textbf{37.1} & 74.8 & 33.1 & 74.0 & 32.4 & 70.6 & 52.5 \\
PedagogicalRL & 93.0 & 21.9 & 70.9 & 18.7 & 73.9 & 52.8 & 76.6 & 43.4 & 80.1 & 63.2 \\
\midrule
\methodname{}\textsubscript{NP} & 92.1 & 21.7 & 72.7 & 32.8 & 75.9 & 65.3 & 82.5 & 64.2 & 81.7 & 73.4 \\
\methodname{}\textsubscript{DP} & \textbf{93.2} & 21.9 & \textbf{73.5} & 30.3 & \textbf{77.0} & \textbf{72.0} & \textbf{90.3} & \textbf{67.3} & \textbf{87.2} & \textbf{79.2} \\
\bottomrule
\end{tabular}
\end{table}

\noindent
\textbf{Results}
Table~\ref{tab:mathtutorbench} shows that the model trained with \methodname{} also performs well on MathTutorBench, especially on teacher response generation. \methodname{}\textsubscript{DP} achieves the highest Pedagogy Average among the evaluated models, at 79.2\%, compared with 52.5\% for the untrained backbone and 63.2\% for PedagogicalRL, and improves on all four constituent metrics. Training raises instruction-following win rates from 74.0\% to 90.3\% in the standard setting and from 70.6\% to 87.2\% in the hard setting. Scaffolding scores also increase from 33.1\% to 72.0\% and from 32.4\% to 67.3\%, respectively. These gains indicate that training improves both adherence to students' instructional requests and the ability to provide guidance that supports their progress. Meanwhile, mistake-location accuracy declines across all trained teachers, and most also regress on Socratic questioning, suggesting a side effect of specific training for teaching. Nonetheless, \methodname{}\textsubscript{DP} retains higher mistake-location accuracy than PedagogicalRL (30.3\% versus 18.7\%).

\begin{figure}[t]
\centering
\begingroup
\definecolor{humanCaseStudent}{HTML}{EDF3F8}
\definecolor{humanCaseTeacher}{HTML}{FFF5DF}
\definecolor{humanCaseBorder}{HTML}{A69A86}
\definecolor{humanCaseAccent}{HTML}{926000}
\setlength{\fboxsep}{7pt}
\setlength{\fboxrule}{0.45pt}
\small
\fcolorbox{humanCaseBorder}{humanCaseStudent}{%
\parbox{\dimexpr\linewidth-2\fboxsep-2\fboxrule\relax}{%
\raggedright
\textbf{Student}\par\smallskip
For $f(x)=\tan(b(x-c))$, the asymptotes are at $0$, $2\pi$, and $4\pi$. I think $b$ is $2\pi$. Could you show me a couple of approaches to finding $b$ and $c$ so I can compare them?}}
\par\vspace{5pt}
\fcolorbox{humanCaseBorder}{humanCaseTeacher}{%
\parbox{\dimexpr\linewidth-2\fboxsep-2\fboxrule\relax}{%
\begin{tabular}{@{}p{\dimexpr0.40\linewidth-7.2pt\relax}@{\hspace{7pt}{\color{humanCaseBorder}\vrule width .4pt}\hspace{7pt}}p{\dimexpr0.60\linewidth-7.2pt\relax}@{}}
\raggedright
\textbf{Base}\par\smallskip
The distance between the asymptotes is $2\pi$, so the period is $2\pi$. For $f(x)=\tan(b(x-c))$, the period is $\pi/(|b|)$. So, $\pi/(|b|)=2\pi$, which means $|b|=1/2$. Therefore, $b=\pm1/2$.
\par\smallskip
To find $c$, note that the asymptotes occur where $b(x-c)=\pi/2+n\pi$. For the first asymptote at $x=0$, we solve $b(-c)=\pi/2$. Using $b=1/2$, we get $-(1/2)c=\pi/2$, so $c=-\pi$.
\par\smallskip
So, $b=1/2$ and $c=-\pi$.
&
\raggedright
\textbf{\textcolor{humanCaseAccent}{\methodname{}\textsubscript{DP}}}\par\smallskip
Sure! Here are two approaches:
\par\smallskip
1. \textbf{Asymptote Distance Approach}: The tangent function $\tan(\theta)$ has asymptotes every $\pi$ units. If the asymptotes of $\tan(b(x-c))$ are spaced $2\pi$ apart, this means the period of the function is $2\pi$. Since the period of $\tan(bx)$ is $\pi/(|b|)$, we can set $\pi/(|b|)=2\pi$ to solve for $b$.
\par\smallskip
2. \textbf{Asymptote Location Approach}: The asymptotes of $\tan(b(x-c))$ occur where $b(x-c)=\pi/2+n\pi$ for integers $n$. Solving for $x$, we get $x=c+\pi/(2b)+n\pi/b$. The asymptotes are at $x=0$, $2\pi$, and $4\pi$, so we can set up equations using these values to solve for $b$ and $c$.
\par\smallskip
Let's start with the first approach.
\end{tabular}}}
\endgroup
\caption{Responding to a student's request to compare approaches. The base model proceeds directly to a solution; \methodname{}\textsubscript{DP} presents two approaches and invites the student to begin with one.}
\label{fig:human-case-contrastive}
\end{figure}

\subsection{Human Pairwise Comparison}
\label{sec:human-pairwise}

\noindent
\textbf{Setup}
To evaluate the effectiveness of our model in real human teaching, we construct 320 dialogue contexts from 80 CoMTA conversations~\citep{miller2024comta} labeled ``Answer Not Accepted''.
\begin{wraptable}{r}{0.6\textwidth}
\vspace{-8pt}
\centering\resulttablesetup
\caption{Human pairwise preference results, with 80 response pairs and
240 judgments for each request type. Win rate excludes ties.}
\label{tab:human-pairwise}
\begin{tabular}{lcccc}
\toprule
Student request & \methodname{}\textsubscript{DP} & Ties & Base & Win rate \\
\midrule
Subgoal decomposition    & 167 & 19 & 54 & 75.6\% \\
Contrastive comparison   & 201 & 5  & 34 & 85.5\% \\
Causal justification     & 185 & 10 & 45 & 80.4\% \\
Error anticipation       & 175 & 11 & 54 & 76.4\% \\
\midrule
\textbf{Overall} &
\textbf{728} & \textbf{45} & \textbf{187} & \textbf{79.6\%} \\
\bottomrule
\end{tabular}
\vspace{-8pt}
\end{wraptable}
For each conversation, we create four variants of the final student response by injecting explicit preferences: two ID preferences (subgoal decomposition and contrastive comparison) and two OOD preferences (causal justification and error anticipation).
For each context, Qwen3-8B and \methodname{}\textsubscript{DP} each generate a short tutor response. We then recruit 96 high school teachers via Prolific~\footnote{\url{https://www.prolific.com/}} to answer ``\emph{Which response teaches the student better?}'' and explain their choice. Each response pair receives three randomized judgments, yielding 960 judgments; ties are allowed.

\noindent
\textbf{Results}
Teachers substantially prefer our model: it receives 728 judgments versus 187 for the base model, corresponding to a 79.6\% win rate excluding ties. The advantage holds across all four preferences (75.6-85.5\%), including the two preferences unseen during training. Moreover, 87 of 96 teachers favor \methodname{}\textsubscript{DP} more often across their judgments, and majority voting prefers it on 270 of 320 response pairs versus 34 for the base model. Figure~\ref{fig:human-case-contrastive} illustrates a representative comparison.
Participants' explanations most often attribute this preference to better guidance tailored to the student's requested approach, and sometimes to better identification and correction of mistakes. A minority prefer the base response when they find our explanations potentially overwhelming, suggesting that improved adaptivity can occasionally come at the cost of simplicity.

\begin{figure}[t]
\centering
\includegraphics[width=\linewidth]{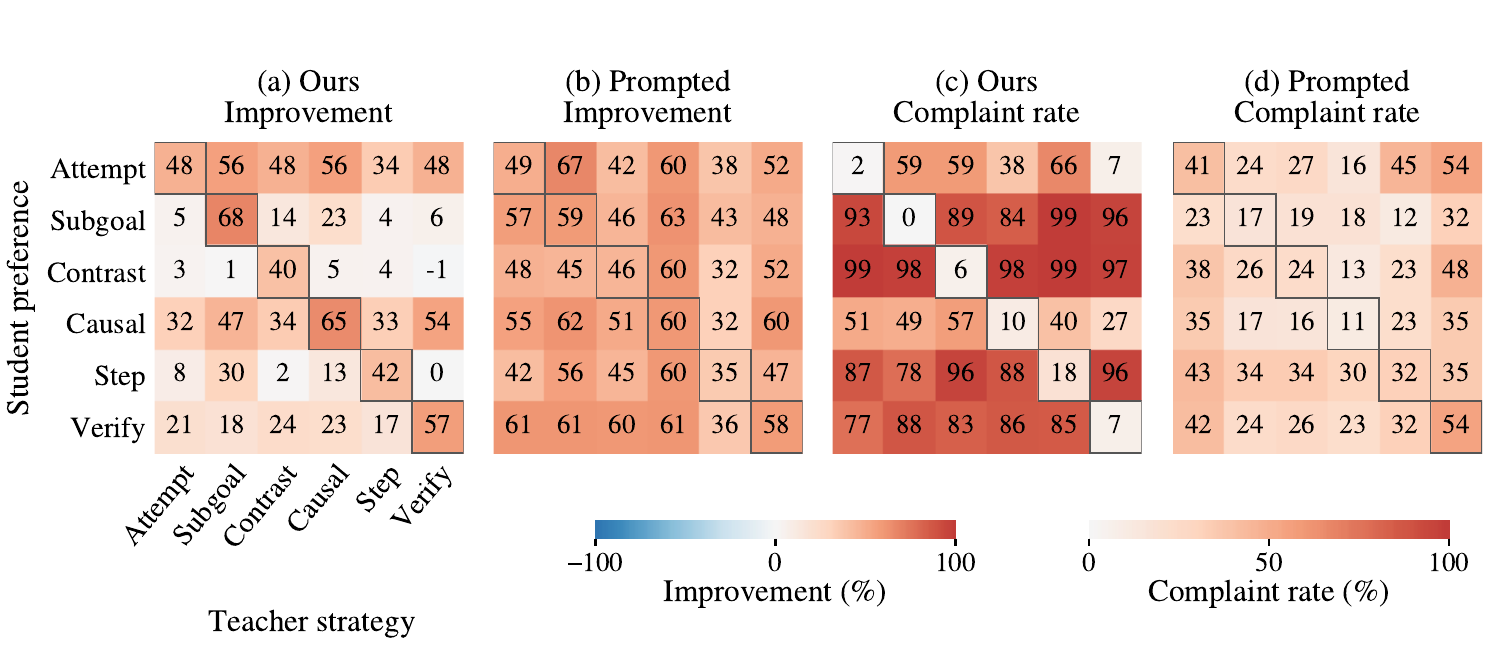}
\caption{Student archetype comparisons between our design and prompted students. Panels (a--b) show student improvement, and panels (c--d) show complaint rates. A good student archetype design should show higher improvement and lower complaint rate over the diagonal, as the students should complain without improvement when the teacher strategy and student preference is unmatched.
}
\label{fig:simulator-comparison}
\end{figure}

\subsection{Evaluating the Training Design}
\label{sec:simulator-ablation}

\noindent
\textbf{Student Archetypes for Teacher Training}
We evaluate and compare our student archetype design with \emph{prompted students}, which are given the same six preferences explicitly in their prompts.
In Figure~\ref{fig:simulator-comparison}, we pair a GPT-5.6-Luna teacher using one of six fixed teaching strategies with a student having one of six preferences, yielding a $6\times6$ matrix for each design; diagonal cells correspond to matched teacher-student pairs. \emph{Selectivity} is measured by differentiated improvement, where a strong design should show higher values along the diagonal. \emph{Informativeness} is measured by differentiated complaint rate: when instruction is unsuitable, students return a prescribed complaint, so matched pairs should exhibit lower complaint rates.
We find that prompted students show neither a clear diagonal advantage in improvement nor consistently lower complaint rates along the diagonal.
On the other hand, our archetypes show both desired properties: improvement is concentrated along the diagonal, while complaint rates are lower for matched pairs.

\begin{wrapfigure}{r}{0.5\textwidth}
\vspace{-8pt}
\centering
\includegraphics[width=\linewidth]{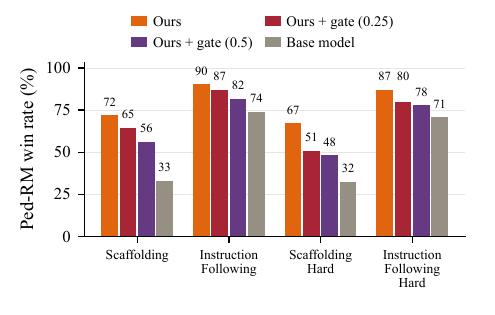}
\caption{Teaching performance under different training objectives. Ped-RM win rates (\%) on the four MathTutorBench response-generation tasks.}
\label{fig:training-dynamics}
\vspace{-8pt}
\end{wrapfigure}

\noindent
\textbf{Training Teachers through Student Improvement}
\methodname{} uses improvements in students' task-solving performance as a training signal for teachers. To examine how additional penalties affect teaching performance, we compare penalty magnitudes of 0.25 and 0.5 for failures of either gate.
The results show that adding gate penalties substantially affects teaching performance. On held-out MathTutorBench (Figure~\ref{fig:training-dynamics}), performance across all four pedagogical dimensions declines as the penalty increases, with Pedagogy Average dropping from 79.2\% without gate penalties to 70.6\% with a penalty of 0.25 and 66.0\% with a penalty of 0.5. We additionally evaluate these variants on the main student archetypes. A small penalty leaves overall performance largely unchanged, whereas a larger penalty markedly reduces performance, including generalization to OOD archetypes. These results suggest that adding penalties based on predefined criteria harms generalization, whereas training based on student improvement allows broader teaching capabilities to emerge. Further experimental details and comparisons with other methods using predefined criteria are provided in Appendix~\ref{sec:pedrl-training-dynamics}.

\section{Related Work}
\label{sec:related-work}

\paragraph{Evaluating LLM teachers.}
Prior work has developed benchmarks and interaction frameworks for evaluating and supporting teaching. MathTutorBench and TutorBench assess pedagogical skills beyond problem solving, revealing limitations in understanding students and providing effective guidance \citep{macina-etal-2025-mathtutorbench,srinivasa2025tutorbench}. LongTutor extends evaluation to diagnosis and adaptive instruction grounded in learning histories \citep{li-etal-2026-longtutor}. System-level approaches introduce pedagogical structure around LLM backbones through explicit strategy steering \citep{puech2024steering}, course planning and reflection \citep{chen2024chattutor}, and instructional suggestions for human tutors \citep{wang2024tutorcopilot}. These studies establish evaluation foundations and demonstrate how structured interactions can support teaching. Our work addresses the complementary problem of training the underlying model to provide adaptive guidance.

\paragraph{Training LLM teachers.}
Supervised approaches develop teaching capabilities from human-authored or synthetic demonstrations, including MathDial, SocraticLM, and LearnLM \citep{macina-etal-2023-mathdial,liu2024socraticlm,learnlm2024}. Preference optimization further shapes teaching behavior through comparisons of candidate responses \citep{sonkar2024alignment,scarlatos2025training}, making supervision dependent on the quality and coverage of collected or generated data. Online reinforcement learning allows teachers to learn from their own interactions with students. PedagogicalRL combines student success with leakage and helpfulness judgments \citep{dinucu-jianu-etal-2025-problem}, while PEARL optimizes multiple explicitly defined pedagogical criteria \citep{chang2026pearl}. These approaches directly specify desirable teacher behavior through scoring rules. We instead treat these qualities as preferences of the learner. This formulation accommodates diverse teaching requirements through students and trains teachers based on student improvement.

\paragraph{Student simulation for teacher training.}
Previous methods model learner responses and errors through prompting or fine-tuning, although mimicking student behavior does not by itself establish responsiveness to teacher instruction \citep{scarlatos2026simulated}. Training on student behavior can also impair a model's knowledge and reasoning \citep{sonkar2024paradox}. Recent work, therefore, explicitly considers both behavioral fidelity and guidance responsiveness \citep{yang2026studentsim}. Another approach uses unlearning to construct novice models and studies their subsequent relearning \citep{song2026novice}. However, controlling knowledge through unlearning remains challenging: targeted information may remain accessible, while retained capabilities can be affected \citep{lynch2024eight,thaker2025unlearning}.
We intentionally build on a less capable student model and encode learner preferences through its adaptive gate, without additional parameter updates to mimic errors or suppress knowledge. Our goal is to construct student archetypes with distinct instructional needs while preserving the underlying model’s ability to benefit from guidance, as human students do. This ensures that improvements in student performance can serve as a reliable training signal for the teacher.

\section{Conclusion}
\label{sec:conclusion}
In this work, to improve LLMs' intrinsic teaching capabilities for diverse learning preferences, we developed \methodname{}, a framework that learns adaptive guidance through interaction with student archetypes. Our framework cleanly separates teacher and student, using gating to instantiate different learning preferences and enable the teacher to learn adaptive guidance.
Instead of picking teaching metrics that capture diverse real-world needs, we train the teacher based on student improvement. Our results show that training improves student outcomes across different archetypes, with gains also extending to an external teaching benchmark. Moreover, human teachers compare our model's responses with those of the base model and judge our model's teaching responses to be substantially better. Further analysis demonstrates the importance of our student archetype design and the effectiveness and generalization benefits of training based on student improvement.

Looking ahead, the teacher-student separation allows us to refine the instructional needs represented by student archetypes independently of teacher training. Our evaluations show that \methodname{} helps train LLM teachers that are more effective at supporting student models and better aligned with human teachers' judgments, taking a further step toward using LLMs to directly support human students in real-world settings. Future work could consider more specific scenarios in which students have individual preferences, including combinations of our archetypes or beyond. Students may also be unaware of their own preferences, requiring teachers to gradually infer them through interaction. We hope this work provides a foundation for such efforts and helps translate the growing knowledge and reasoning capabilities of LLMs into teaching that strengthens humans' understanding.

\subsection*{AI Use Statement}
During the research process, we used Gemini to generate student-message continuations for evaluation and AI tools to organize data. During manuscript preparation, AI tools helped identify relevant literature, search for information, and prepare tables and figures. All AI-assisted content was reviewed by the authors, who take responsibility for the final content of this work.

\subsection*{Ethics Statement}
Our human evaluation recruited participants through Prolific. All participants provided informed consent and were compensated for their time. Participant privacy was protected through Prolific's privacy safeguards.

\subsection*{Reproducibility Statement}
We describe our student archetypes, gating mechanisms, reward construction, and optimization procedure in the main text and appendix. The appendix further provides training and generation settings, evaluation protocols, prompts, and details of our asynchronous implementation using AReaL. These descriptions support reproduction of our training and evaluation procedures.

\bibliography{references}
\bibliographystyle{iclr2027_conference}

\clearpage
\appendix
\section{Objective and Derivation}
\label{app:masked-objective}

\subsection{Standard Formulation}
\label{app:standard-formulation}

We optimize the following policy-gradient surrogate with the masked return used in Eq.~\ref{eq:turn-advantage}. Let $\pi_{\theta_{\mathrm{old}}}$ denote the rollout policy and
\[
\rho_{i,t}(\theta)
=
\frac{
\pi_{\theta}(y_{i,t}\mid C_{i,t}^{T})
}{
\pi_{\theta_{\mathrm{old}}}(y_{i,t}\mid C_{i,t}^{T})
}
\]
denote the likelihood ratio for the teacher action at turn $t$. Before centering and normalization, consider the masked policy-gradient surrogate
\begin{equation}
\mathcal L_{\mathrm{mask}}(\theta;\theta_{\mathrm{old}})
=
\mathbb E_{\tau\sim\pi_{\theta_{\mathrm{old}}}}
\left[
\sum_t
\rho_t(\theta)\,
m_t R
\right].
\label{eq:masked-surrogate}
\end{equation}
At the rollout policy, $\rho_t(\theta_{\mathrm{old}})=1$, and
\begin{equation}
\left.
\nabla_\theta
\mathcal L_{\mathrm{mask}}(\theta;\theta_{\mathrm{old}})
\right|_{\theta=\theta_{\mathrm{old}}}
=
\mathbb E_{\tau\sim\pi_{\theta_{\mathrm{old}}}}
\left[
\sum_t
m_t R\,
\nabla_\theta
\log
\pi_\theta(y_t\mid C_t^{T})
\right]_{\theta=\theta_{\mathrm{old}}}.
\label{eq:masked-gradient}
\end{equation}
Thus, the masked return $U_t=m_tR$ serves as the return weight for the corresponding turn-level policy-gradient term in this surrogate. In particular, masking operates on the return rather than directly on the policy update: an invalid or gate-rejected turn has $U_t=0$, while its final advantage may still be nonzero after leave-one-out centering, normalization, and the addition of local penalties.

In the standard formulation, we construct the turn-level advantage $A_{i,t}$ according to Eq.~\ref{eq:turn-advantage} and broadcast it to all teacher-generated tokens in turn $t$. Let $k$ index these tokens, and let $C_{i,t,k}^{T}$ contain $C_{i,t}^{T}$ followed by the preceding tokens $y_{i,t,<k}$ of the current response. We define
\[
\rho_{i,t,k}(\theta)
=
\frac{
\pi_\theta(y_{i,t,k}\mid C_{i,t,k}^{T})
}{
\pi_{\theta_{\mathrm{old}}}(y_{i,t,k}\mid C_{i,t,k}^{T})
}.
\]
We then optimize the standard PPO clipped surrogate
\begin{equation}
\mathcal L_{\mathrm{PPO}}(\theta)
=
\mathbb E
\left[
\sum_{i,t,k}
\min\left(
\rho_{i,t,k}(\theta) A_{i,t},
\operatorname{clip}
\left(
\rho_{i,t,k}(\theta),
1-\epsilon_{\mathrm{clip}},
1+\epsilon_{\mathrm{clip}}
\right)
A_{i,t}
\right)
\right].
\label{eq:masked-ppo}
\end{equation}
Leave-one-out centering and batch normalization are applied to the masked returns before constructing $A_{i,t}$, as described in the main paper. The return mask therefore determines which turns receive the trajectory-level outcome return, while the subsequent operations, including PPO clipping, follow the standard optimization pipeline.

\subsection{Asynchronous Implementation}
\label{app:async-implementation}

We use asynchronous training in AReaL~\citep{fu2025areal} to overlap rollout generation with policy optimization and reduce GPU idle time. Unlike the standard formulation in Appendix~\ref{app:standard-formulation}, rollout samples may come from older policies. We therefore use decoupled PPO to separate behavior-policy correction from the proximal policy used for PPO clipping.

Let $\mu_{i,t,k}$ denote the behavior policy under which token $y_{i,t,k}$ was sampled, and let $\pi_{\theta_{\mathrm{prox}}}$ denote the teacher policy immediately before the current training update. We define
\begin{equation}
\begin{aligned}
r_{i,t,k}(\theta)
&=
\frac{\pi_\theta(y_{i,t,k}\mid C_{i,t,k}^{T})}
{\pi_{\theta_{\mathrm{prox}}}(y_{i,t,k}\mid C_{i,t,k}^{T})},
\qquad
w_{i,t,k}
=
\frac{\pi_{\theta_{\mathrm{prox}}}(y_{i,t,k}\mid C_{i,t,k}^{T})}
{\mu_{i,t,k}(y_{i,t,k}\mid C_{i,t,k}^{T})}.
\end{aligned}
\label{eq:async-ratios}
\end{equation}

Here, $w_{i,t,k}$ accounts for the off-policy correction, while clipping $r_{i,t,k}$ constrains updates relative to the recent proximal policy rather than an outdated rollout policy. The decoupled objective is
\begin{equation}
\mathcal J_{\mathrm{dec}}(\theta)
=
\widehat{\mathbb E}_{\mathcal T}
\left[
w_{i,t,k}
\min\!\left(
r_{i,t,k}(\theta)A_{i,t},
\operatorname{clip}\!\left(
r_{i,t,k}(\theta),
1-\epsilon_{\mathrm{clip}},
1+\epsilon_{\mathrm{clip}}
\right)A_{i,t}
\right)
\right],
\label{eq:async-ppo}
\end{equation}
where $\widehat{\mathbb E}_{\mathcal T}$ averages over teacher-generated training tokens, with advantages and importance weights treated as constant. In the synchronous case where the behavior and proximal policies coincide, $w_{i,t,k}=1$, and the per-token objective reduces to the standard PPO clipped surrogate in Appendix~\ref{app:standard-formulation}.

For additional stability, our implementation uses dual clipping with coefficient $c=3$ and importance-weight masking with threshold $\kappa=5$; these safeguards take effect only when their respective bounds are exceeded.

\section{Experimental Setup and Details}
\label{sec:results-evaluation-details}

\subsection{Training and Generation Settings}
\label{sec:training-generation-settings}
\paragraph{Data and models.}
We use the 759 training and 528 evaluation problems retained by the MATH filtering procedure described in \S\ref{sec:experimental-setup}.
During training, we disable thinking for both the Qwen3-8B teacher and the Qwen3-1.7B student and use Qwen3-8B as the LLM judge for the guidance gate, adaptive gate, and answer judge. \methodname{}\textsubscript{DP} trains on the four ID archetypes (None, Attempt, Subgoal, and Contrast); \methodname{}\textsubscript{NP} trains on a single student archetype without a teaching strategy requirement (None). We also use a single student with other ID archetypes to train teachers and report their results in Appendix~\ref{sec:single-archetype-results}.

\paragraph{Optimization.}
We implement training on AReaL~\citep{fu2025areal}, overlapping rollout generation with policy optimization, with up to 64 concurrent rollout tasks, a consumer batch size of 16, and a maximum policy-version lag of two.
The teacher uses rank-16 LoRA adapters with scaling parameter 16 on all linear projections. Each training step draws 16 problems and samples eight trajectories per problem. We train for 1500 updates using Adam with a learning rate of $5\times10^{-5}$, coefficients $(0.9,0.999)$, epsilon $10^{-8}$, no weight decay, gradient clipping at $1.0$, and a linear learning-rate warmup over one epoch, followed by a constant learning rate. %
 We use a turn-level leave-one-out group baseline, a policy clipping threshold of 0.2, and no KL penalty. The student-improvement reward is assigned only to valid turns that pass both gates. Malformed outputs and exact repeated teacher responses receive a separate format or repetition penalty of $-0.5$.

\paragraph{Preparation, tutoring, and testing.}
During training, the teacher has up to three preparation attempts, each capped at $4096$ tokens. A verified correct solution is kept private to the teacher and shared across trajectories for the same problem; a problem is skipped if no attempt succeeds. Tutoring permits at most $10$ turns, and the teacher may end earlier. During testing, we sample $8$ unaided and $8$ post-tutoring samples on the original problem to estimate the student's improvement. Teacher sampling uses temperature 1.0 and top-$p$ 1.0, with a cap of $1024$ tokens per tutoring response. The training prompts and response format are provided in Appendix~\ref{sec:training-prompts}.

\subsection{Evaluation on Student Archetypes}
\label{sec:archetype-evaluation-details}

\paragraph{Simulator evaluation settings.}
Every teacher is evaluated on all $528$ problems and seven archetypes.
The $10$-turn tutoring budget and $8$ testing samples stay the same as training.
To make the comparison fairer, preparation is enabled without correctness verification or selection.
Teachers with local LLMs use temperature $1.0$, top-$p$ $1.0$, and at most $2048$ tokens per turn with thinking disabled.
Teachers with Gemini 3.8 Flash use the default setting: \texttt{medium} reasoning with temperature $1.0$.
The student uses temperature $0.7$, top-$p$ $0.8$, top-$k$ $20$, and a $2048$-token cap.
To make the comparison fairer and more reliable, we use Qwen3.8-27B-FP8 for gating and answer judgments, with temperature $0$ and thinking disabled.
Teacher-private reasoning is not exposed to the student.
Since untrained Qwen3-8B struggles to follow the required format in long contexts, we remove the private-reasoning format requirement and let it output the response directly. Appendix~\ref{sec:archetype-evaluation-prompts} specifies the response templates and answer judge.

\paragraph{PedagogicalRL.}
We implement the PedagogicalRL baseline on the AReaL infrastructure following its official method, which combines student success with leakage and helpfulness judgments. The baseline uses the same Qwen3-8B backbone, rank-16 LoRA adapters, and 1500 training updates. Its frozen configuration uses eight sampled trajectories per group, a $5\times10^{-5}$ learning rate, a warmup proportion of $0.001$, and a $1024$-token response cap. Its evaluation uses the same student archetypes and the held-out judge as our teachers do. We align its required output format with ours so that subsequent comparisons are not affected by differences in format requirements.

\paragraph{Uncertainty estimates.}
We compute 95\% percentile confidence intervals using 10000 question-level bootstrap resamples. Each resample draws 528 problems with replacement, retaining all seven student archetypes and their eight test outcomes per problem, together with the corresponding unaided baselines. The same sampled problem indices are used across teachers for paired comparisons. These intervals quantify question-sampling uncertainty.

\subsection{MathTutorBench Evaluation Details}
\label{sec:mathtutorbench-evaluation-details}

\paragraph{Measurement procedures.}
Problem Solving and Mistake Correction are measured by accuracy, Socratic Questioning by BLEU against reference questions, Solution Correctness by F1 for classifying student solutions as correct or incorrect, and Mistake Location by micro-F1 for identifying the first incorrect step. The four teacher response generation tasks use Ped-RM win rates: the proportion of model responses preferred over reference human-teacher responses by the benchmark's pedagogical reward model. Pedagogy Average is the mean of these four win rates.

\paragraph{Response processing.}
We retain the benchmark task prompts and apply the following response-processing rules. Because the original benchmark stopping rules can unnecessarily truncate evaluated responses at paragraph breaks or task headings, we keep only the line-start Teacher and Student labels as stopping boundaries and remove the other stopping rules. For Solution Correctness, we use the final explicit Yes/No judgment rather than the first, since the model may provide its answer after reasoning.

\paragraph{Generation settings.}
Following the benchmark's default decoding settings, Qwen models use temperature $0$ and a $2048$-token output cap, with thinking disabled. Gemini uses \texttt{medium} reasoning with temperature $1.0$.

\subsection{Human Study Details}
\label{sec:human-study-details}

\paragraph{Dialogue construction.}
We use 80 CoMTA dialogue contexts labeled \texttt{Answer Not Accepted}. For each context, Gemini generates one short student request for each of four instructional needs: Subgoal, Contrast, Causal, and Error Anticipation. The generated request is appended to the student's original wording. The prompt for generating these requests is provided in Appendix~\ref{sec:human-study-prompts}. Subgoal and Contrast correspond to training preferences, while Causal and Error Anticipation are unseen instructional needs. Because not all preferences can be naturally appended to the student's response, we introduce Error Anticipation as a new held-out preference.

\paragraph{Response generation and presentation.}
We compare \methodname{}\textsubscript{DP} with the untrained Qwen3-8B backbone. Both receive the same tutoring prompt, which requests a short response and asks the teacher to check and, where necessary, correct the student's last message. Generation uses temperature $0.7$, top-$p$ $0.8$, top-$k$ $20$, a $2048$-token cap, and disabled thinking. Only the student-facing output is displayed, with private reasoning removed and mathematical notation converted to readable plain text.
Model identities are hidden, and order is randomized during comparison. The generation and annotation prompts are in Appendix~\ref{sec:human-study-prompts}.

\paragraph{Annotation and aggregation.}
The full study recruited 96 high school teachers with at least a bachelor's degree and English proficiency via Prolific. Each teacher evaluated ten items, with three judgments per item, for 960 judgments. Participants read the conversation and both responses, selected ``\texttt{Response A}'', ``\texttt{Response B}'', or ``\texttt{About the same}'', and optionally provided a brief explanation. Instructions ask them to compare teaching quality and not judge by length, formatting, or other surface features alone. Compensation rate was \$16 per hour. The final dataset contains three judgments for every item.

\subsection{Student Simulator Analysis Settings}
\label{sec:simulator-analysis-details}

\paragraph{Shared design and sampling.}
We sample 50 problems without replacement from the 528-problem evaluation set to conduct the analysis. The teacher is GPT-5.6-Luna with medium reasoning, with instruction to teach using one assigned strategy throughout the dialogue (the prompts are provided in Appendix~\ref{sec:simulator-prompts}).
Other evaluation setting is the same as default.

\paragraph{Simulator variants.}
For our student archetypes, when the gate rejects a message, we randomly select one of six fixed responses that only express the student's lack of understanding, avoiding the influence of explanations of the rejection on teaching.
The prompted variant disables external adaptive gating and instead appends the same preference criterion to each student request.
It asks the student to assess the teacher's latest message internally, work on the mathematics if the criterion is met, and otherwise return only a specified complaint from the same pool.

\subsection{Training-objective Comparison Settings}
\label{sec:training-dynamics-details}
We compare the base Qwen3-8B model, \methodname{}\textsubscript{DP}, and two gate-penalty variants with penalty magnitudes of 0.25 and 0.5. Both variants use the same training setup as \methodname{}\textsubscript{DP}. In each variant, we apply penalties for gate violations, with a magnitude of 0.25 or 0.5 depending on the variant.

\section{Additional Results and Analyses}
\label{sec:additional-results}

\subsection{Effect of Preparation}
\label{sec:preparation-evaluation}

\begin{table}[!htb]
\centering\resulttablesetup
\caption{Student test accuracy after teaching (\%) by Qwen3-8B with and without preparation. Overall 95\% confidence intervals use the same question-level bootstrap procedure as the main table.}
\label{tab:preparation-evaluation}
\begin{tabular}{lccccccccccc}
\toprule
& \multicolumn{4}{c}{ID} & \multicolumn{3}{c}{OOD} & \multicolumn{3}{c}{Average} & \\
\cmidrule(lr){2-5}\cmidrule(lr){6-8}\cmidrule(lr){9-11}
Preparation & \prefNone & \prefAtt & \prefSub & \prefCtr & \prefCau & \prefStp & \prefVer & ID & OOD & \prefOvl & 95\% CI \\
\midrule
Without & 45.0 & 44.3 & 44.6 & 45.7 & 49.8 & 44.3 & 43.5 & 44.9 & 45.9 & 45.3 & [42.9, 47.8] \\
With & 53.2 & 50.7 & 48.4 & 44.1 & 54.4 & 47.8 & 41.5 & 49.1 & 47.9 & 48.6 & [46.1, 51.0] \\
\bottomrule
\end{tabular}
\end{table}

To examine the effect of preparation on teaching performance, we evaluate Qwen3-8B with and without preparation, using the default evaluation setting.
In Table \ref{tab:preparation-evaluation}, preparation raises overall accuracy from 45.3\% to 48.6\%, a gain of 3.3 percentage points. ID accuracy increases from 44.9\% to 49.1\%, while OOD accuracy increases from 45.9\% to 47.9\%.
These results support the use of preparation in the main evaluation, while showing that its benefit is not uniform across preferences.

\subsection{Additional Evaluation on Student Archetypes}
\label{sec:additional-archetype-evaluation}

\subsubsection{Training on Individual Archetypes}
\label{sec:single-archetype-results}
To ablate the student archetypes used for training, we compare teachers trained on individual archetypes with the teacher trained on all ID archetypes.
We include teachers trained with a single student with Attempt, Subgoal, and Contrast as preferences in Table~\ref{tab:single-archetype-results}, using the same training and evaluation setting as \methodname{}\textsubscript{DP}.
We additionally include Gemini 3.8 Flash as a strong external general-purpose model in the complete results below. It is reported separately from the Qwen3-8B models and excluded from the bold highlighting.

\begin{table}[!htb]
\centering\resulttablesetup
\caption{Complete student-archetype results, including all single-archetype teachers and Gemini 3.8 Flash as an external reference. Values are student test accuracy after teaching (\%). Bold values indicate the best result in each column among models sharing the Qwen3-8B backbone.}
\label{tab:single-archetype-results}
\begin{tabular}{lcccccccccc}
\toprule
Teacher & \prefNone & \prefAtt & \prefSub & \prefCtr & \prefCau & \prefStp & \prefVer & ID & OOD & \prefOvl \\
\midrule
Qwen3-8B & 53.2 & 50.7 & 48.4 & 44.1 & 54.4 & 47.8 & 41.5 & 49.1 & 47.9 & 48.6 \\
PedagogicalRL & 63.0 & 59.4 & 54.7 & 44.4 & 59.0 & 44.3 & 43.8 & 55.4 & 49.1 & 52.7 \\
\midrule
\methodname{}\textsubscript{NP} & 63.1 & 58.8 & 52.0 & 45.3 & 55.1 & 40.6 & 49.2 & 54.8 & 48.3 & 52.0 \\
\methodname{}\textsubscript{Attempt} & 73.1 & 72.5 & 62.5 & 53.9 & 63.1 & 46.2 & 57.0 & 65.5 & 55.4 & 61.2 \\
\methodname{}\textsubscript{Subgoal} & \textbf{78.7} & \textbf{79.2} & \textbf{75.1} & \textbf{65.0} & 73.1 & 40.8 & 65.8 & \textbf{74.5} & 59.9 & 68.2 \\
\methodname{}\textsubscript{Contrast} & 64.9 & 66.7 & 62.5 & 57.5 & 59.8 & 46.0 & 59.5 & 62.9 & 55.1 & 59.6 \\
\methodname{}\textsubscript{DP} & 76.2 & 78.2 & 72.7 & 60.7 & \textbf{77.1} & \textbf{51.8} & \textbf{66.4} & 72.0 & \textbf{65.1} & \textbf{69.0} \\
\midrule
Gemini 3.8 Flash & 78.1 & 79.2 & 74.4 & 68.4 & 72.8 & 61.6 & 55.8 & 75.0 & 63.4 & 70.1 \\
\bottomrule
\end{tabular}
\end{table}

Table~\ref{tab:single-archetype-results} shows that training the Qwen3-8B base model with \methodname{} enables \methodname{}\textsubscript{DP} to achieve overall student test accuracy comparable to Gemini 3.8 Flash (69.0\% versus 70.1\%), and each single-archetype teacher exceeds \methodname{}\textsubscript{NP}'s overall accuracy of 52.0\%. However, training on a preference does not consistently yield the best performance on that same preference. Among the Qwen3-8B teachers, Subgoal has the highest ID average, at 74.5\%, while \methodname{}\textsubscript{DP} achieves the highest OOD and overall averages, at 65.1\% and 69.0\%, respectively. Although the Subgoal-trained teacher also performs well, we argue that this is because the Subgoal preference itself is particularly helpful to the student model.
Table~\ref{tab:single-archetype-mtb} further demonstrate that training on all ID archetypes yields the strongest overall teaching performance and generalization on MathTutorBench.

\begin{table}[!htb]
\centering\resulttablesetup
\caption{Complete MathTutorBench scores (\%), including all single-archetype teachers and Gemini 3.8 Flash as an external reference. Bold values indicate the best result in each column among models sharing the Qwen3-8B backbone.}
\label{tab:single-archetype-mtb}
\begin{tabular}{lcccccccccc}
\toprule
Teacher & PS & SQ & SC & ML & MC & Scaf. & IF & Scaf.-H & IF-H & Ped.\ Avg. \\
\midrule
Qwen3-8B & 92.1 & 25.3 & 70.6 & \textbf{37.1} & 74.8 & 33.1 & 74.0 & 32.4 & 70.6 & 52.5 \\
PedagogicalRL & 93.0 & 21.9 & 70.9 & 18.7 & 73.9 & 52.8 & 76.6 & 43.4 & 80.1 & 63.2 \\
\midrule
\methodname{}\textsubscript{NP} & 92.1 & 21.7 & 72.7 & 32.8 & 75.9 & 65.3 & 82.5 & 64.2 & 81.7 & 73.4 \\
\methodname{}\textsubscript{Attempt} & 92.0 & \textbf{27.7} & \textbf{79.1} & 27.5 & 74.0 & 49.0 & 71.9 & 38.5 & 70.0 & 57.4 \\
\methodname{}\textsubscript{Subgoal} & 92.9 & 24.7 & 76.0 & 17.6 & \textbf{77.0} & 62.8 & 80.1 & 51.4 & 73.7 & 67.0 \\
\methodname{}\textsubscript{Contrast} & 92.5 & 24.4 & 73.2 & 34.7 & 70.5 & 65.8 & 88.2 & 55.7 & 81.7 & 72.8 \\
\methodname{}\textsubscript{DP} & \textbf{93.2} & 21.9 & 73.5 & 30.3 & \textbf{77.0} & \textbf{72.0} & \textbf{90.3} & \textbf{67.3} & \textbf{87.2} & \textbf{79.2} \\
\midrule
Gemini 3.8 Flash & 97.6 & 34.8 & 88.9 & 80.3 & 95.7 & 72.1 & 69.9 & 73.4 & 74.3 & 72.4 \\
\bottomrule
\end{tabular}
\end{table}

Different single-archetype teachers exhibit different strengths. While their performance on math expertise and student understanding is broadly comparable, \methodname{}\textsubscript{DP} outperforms all single-archetype teachers on every teacher response generation metric. This demonstrates the improved generalization and teaching capabilities obtained by training with diverse students. As a strong general-purpose model, Gemini scores markedly higher on MathTutorBench's mathematical problem-solving and diagnosis tasks. However, \methodname{}\textsubscript{DP} achieves a higher Pedagogy Average (79.2\% versus 72.4\%) and higher scores on the benchmark's standard and hard instruction-following tasks. Although our base model scores lower than Gemini on mathematical problem solving, training raises its instruction-following win rate from 74.0\% to 90.3\%, exceeding Gemini's 69.9\%, with gains also extending to the hard setting and its overall Pedagogy Average increasing from 52.5\% to 79.2\%. This echoes our initial motivation: stronger problem-solving ability does not automatically translate into better teaching.

\subsubsection{Further Analysis of Teaching Gains}
\label{sec:post-feedback-details}
We examine each teacher's improvement from two perspectives: how often its messages satisfy student preferences and pass the gates, and the teaching quality of those accepted messages.

\noindent\textbf{Satisfying Student Preferences.}
We consider a message to satisfy student preferences when it is valid and passes both the guidance and adaptive gates. For each trajectory, we compute the fraction of teacher messages for interaction that satisfy these conditions and reach the student. Since trajectories differ in dialogue length, we average these fractions across trajectories with equal weights.

\begin{table}[!htb]
\centering\resulttablesetup
\caption{Student preference satisfaction rates (\%).}
\label{tab:post-feedback-acceptance}
\begin{tabular}{lcccccccccc}
\toprule
& \multicolumn{4}{c}{ID} & \multicolumn{3}{c}{OOD} & \multicolumn{3}{c}{Average} \\
\cmidrule(lr){2-5}\cmidrule(lr){6-8}\cmidrule(lr){9-11}
Teacher & \prefNone & \prefAtt & \prefSub & \prefCtr & \prefCau & \prefStp & \prefVer & ID & OOD & \prefOvl \\
\midrule
Qwen3-8B & 83.0 & 53.3 & 44.4 & 24.5 & 38.5 & 25.7 & 43.0 & 51.3 & 35.7 & 44.6 \\
\methodname{}\textsubscript{NP} & 79.4 & 33.9 & 32.9 & 16.9 & 26.9 & 10.7 & 29.8 & 40.8 & 22.5 & 32.9 \\
\methodname{}\textsubscript{Attempt} & 86.4 & 82.7 & 41.6 & 29.8 & 43.2 & 22.7 & 53.0 & 60.1 & 39.6 & 51.3 \\
\methodname{}\textsubscript{Subgoal} & 86.3 & 82.0 & 63.0 & 34.6 & 46.4 & 7.7 & 63.3 & 66.5 & 39.1 & 54.7 \\
\methodname{}\textsubscript{Contrast} & 74.4 & 73.2 & 47.3 & 31.9 & 49.3 & 17.0 & 62.1 & 56.7 & 42.8 & 50.7 \\
\methodname{}\textsubscript{DP} & 82.5 & 81.4 & 57.3 & 36.1 & 60.9 & 21.3 & 62.6 & 64.3 & 48.2 & 57.4 \\
\bottomrule
\end{tabular}
\end{table}

\noindent\textbf{Teaching Quality.}
We examine accepted teaching from two angles: how much a single round helps the student and how many rounds the teacher delivers.

\noindent\textit{Single-round effect.}
We rerun the evaluation, stopping each dialogue immediately after the first accepted teacher message and the student's reply. We report student improvement on trajectories with an accepted message, so that each evaluated student receives exactly one round of teaching.

\begin{table}[!htb]
\centering\resulttablesetup
\caption{Student improvement after one accepted teaching round (pp).}
\label{tab:first-accepted-improvement}
\begin{tabular}{lcccccccccc}
\toprule
& \multicolumn{4}{c}{ID} & \multicolumn{3}{c}{OOD} & \multicolumn{3}{c}{Average} \\
\cmidrule(lr){2-5}\cmidrule(lr){6-8}\cmidrule(lr){9-11}
Teacher & \prefNone & \prefAtt & \prefSub & \prefCtr & \prefCau & \prefStp & \prefVer & ID & OOD & \prefOvl \\
\midrule
Qwen3-8B & -0.7 & -2.3 & 9.9 & 8.9 & 9.9 & 4.9 & -1.8 & 4.0 & 4.3 & 4.1 \\
\methodname{}\textsubscript{NP} & 12.8 & 11.8 & 10.6 & 10.1 & 11.9 & 5.8 & 11.1 & 11.3 & 9.6 & 10.6 \\
\methodname{}\textsubscript{Attempt} & 10.2 & 10.7 & 13.6 & 11.4 & 16.4 & 5.9 & 10.3 & 11.5 & 10.9 & 11.2 \\
\methodname{}\textsubscript{Subgoal} & 28.6 & 27.0 & 23.5 & 23.5 & 23.2 & 9.0 & 29.2 & 25.7 & 20.5 & 23.4 \\
\methodname{}\textsubscript{Contrast} & 23.4 & 27.7 & 25.5 & 26.1 & 32.5 & 13.5 & 28.7 & 25.7 & 24.9 & 25.3 \\
\methodname{}\textsubscript{DP} & 28.1 & 28.9 & 26.1 & 22.7 & 26.8 & 13.9 & 26.6 & 26.4 & 22.4 & 24.7 \\
\bottomrule
\end{tabular}
\end{table}

\noindent\textit{Amount of teaching.}
In the full evaluation, we count the accepted teacher messages per trajectory and report the average number of effective teaching turns received by the student.

\begin{table}[!htb]
\centering\resulttablesetup
\caption{Average number of effective teaching turns.}
\label{tab:effective-teaching-turns}
\begin{tabular}{lcccccccccc}
\toprule
& \multicolumn{4}{c}{ID} & \multicolumn{3}{c}{OOD} & \multicolumn{3}{c}{Average} \\
\cmidrule(lr){2-5}\cmidrule(lr){6-8}\cmidrule(lr){9-11}
Teacher & \prefNone & \prefAtt & \prefSub & \prefCtr & \prefCau & \prefStp & \prefVer & ID & OOD & \prefOvl \\
\midrule
Qwen3-8B & 4.0 & 3.4 & 2.4 & 1.7 & 2.4 & 1.9 & 2.3 & 2.9 & 2.2 & 2.6 \\
\methodname{}\textsubscript{NP} & 6.4 & 1.6 & 2.1 & 1.0 & 1.6 & 0.6 & 1.1 & 2.8 & 1.1 & 2.1 \\
\methodname{}\textsubscript{Attempt} & 2.1 & 1.9 & 1.9 & 1.4 & 1.7 & 1.7 & 1.2 & 1.8 & 1.5 & 1.7 \\
\methodname{}\textsubscript{Subgoal} & 1.6 & 1.5 & 1.6 & 1.5 & 1.3 & 0.6 & 1.0 & 1.6 & 1.0 & 1.3 \\
\methodname{}\textsubscript{Contrast} & 1.1 & 1.1 & 1.0 & 0.9 & 0.9 & 0.8 & 0.9 & 1.0 & 0.9 & 1.0 \\
\methodname{}\textsubscript{DP} & 1.7 & 1.5 & 1.7 & 1.5 & 1.4 & 1.1 & 1.1 & 1.6 & 1.2 & 1.4 \\
\bottomrule
\end{tabular}
\end{table}

\noindent\textbf{Analysis.}
Table~\ref{tab:teaching-gains-summary} summarizes these measurements alongside overall student improvement, averaged across the seven student archetypes.

\begin{table}[!htb]
\centering\resulttablesetup
\caption{Preference satisfaction, teaching quality, and overall student improvement.}
\label{tab:teaching-gains-summary}
\begin{tabular}{lcccc}
\toprule
Teacher & Gate pass rate (\%) & First reply (pp) & Effective turns & Improvement (pp) \\
\midrule
Qwen3-8B & 44.6 & 4.1 & 2.6 & 12.4 \\
\methodname{}\textsubscript{NP} & 32.9 & 10.6 & 2.1 & 16.6 \\
\methodname{}\textsubscript{Attempt} & 51.3 & 11.2 & 1.7 & 25.3 \\
\methodname{}\textsubscript{Subgoal} & 54.7 & 23.4 & 1.3 & 32.0 \\
\methodname{}\textsubscript{Contrast} & 50.7 & 25.3 & 1.0 & 23.9 \\
\methodname{}\textsubscript{DP} & 57.4 & 24.7 & 1.4 & 33.1 \\
\bottomrule
\end{tabular}
\end{table}

\methodname{}\textsubscript{DP} combines the highest preference satisfaction rate with strong single-round teaching, while maintaining a moderate number of effective turns. Compared with the base model and \methodname{}\textsubscript{NP}, it achieves greater overall improvement with fewer accepted turns, suggesting more efficient teaching. It also delivers more effective turns than Subgoal and Contrast while retaining a comparable single-round effect. Contrast illustrates why strong individual replies alone may be insufficient: despite having the highest single-round improvement and a preference satisfaction rate close to Attempt's, it delivers only 1.0 effective turns per trajectory, compared with Attempt's 1.7, and achieves lower overall improvement. These results suggest that teaching gains depend jointly on satisfying student preferences, delivering useful instruction, and sustaining the interaction.

\subsubsection{Generalization to Unseen Student Backbones}
\label{sec:student-backbone-stability}
To examine transfer across student model families, we evaluate the teacher checkpoints used with Gemma-3-1B-IT \citep{team2025gemma} and Llama-3.1-8B-Instruct \citep{grattafiori2024llama} as students.
Table~\ref{tab:cross-student-backbones} shows that \methodname{}\textsubscript{DP} achieves the highest accuracy for every archetype with both student backbones. Overall accuracy rises from 14.3\% to 21.7\% with Gemma and from 37.5\% to 53.5\% with Llama, compared with the untrained teacher. These gains indicate that the learned teaching behavior transfers beyond the Qwen3-1.7B student used during training.
More interestingly, although PedagogicalRL and \methodname{}\textsubscript{NP} improve over the base model in Table~\ref{tab:retest-accuracy}, they provide little or no improvement over the base model when evaluated with other student models, indicating weaker generalization.This further highlights that training with diverse student archetypes generalizes not only to unseen preferences but also to other student models.

\begin{table}[!htb]
\centering\resulttablesetup
\caption{Student test accuracy after teaching (\%) with alternative student backbones. Bold values mark the best teacher for each student backbone. Overall 95\% confidence intervals use 10000 bootstrap resamples of problems.}
\label{tab:cross-student-backbones}
\begin{tabular}{lccccccccccc}
\toprule
& \multicolumn{4}{c}{ID} & \multicolumn{3}{c}{OOD} & \multicolumn{3}{c}{Average} & \\
\cmidrule(lr){2-5}\cmidrule(lr){6-8}\cmidrule(lr){9-11}
Teacher & \prefNone & \prefAtt & \prefSub & \prefCtr & \prefCau & \prefStp & \prefVer & ID & OOD & \prefOvl & 95\% CI \\
\midrule
\multicolumn{12}{c}{Student: Gemma-3-1B-IT} \\
\midrule
Qwen3-8B & 15.2 & 12.3 & 16.0 & 13.6 & 16.4 & 16.0 & 10.7 & 14.3 & 14.4 & 14.3 & [12.7, 16.0] \\
PedagogicalRL & 16.7 & 14.2 & 14.2 & 12.9 & 14.8 & 14.2 & 11.2 & 14.5 & 13.4 & 14.0 & [12.5, 15.6] \\
\methodname{}\textsubscript{NP} & 13.0 & 14.2 & 12.3 & 12.5 & 14.3 & 12.8 & 12.7 & 13.0 & 13.3 & 13.1 & [11.5, 14.8] \\
\methodname{}\textsubscript{DP} & \textbf{25.1} & \textbf{25.6} & \textbf{21.4} & \textbf{17.1} & \textbf{22.3} & \textbf{18.9} & \textbf{21.4} & \textbf{22.3} & \textbf{20.9} & \textbf{21.7} & [19.8, 23.7] \\
\midrule
\multicolumn{12}{c}{Student: Llama-3.1-8B-Instruct} \\
\midrule
Qwen3-8B & 43.1 & 42.6 & 41.9 & 30.7 & 42.5 & 32.7 & 29.0 & 39.6 & 34.7 & 37.5 & [35.1, 40.0] \\
PedagogicalRL & 50.0 & 49.8 & 36.6 & 33.2 & 39.5 & 31.6 & 28.5 & 42.4 & 33.2 & 38.5 & [36.1, 40.9] \\
\methodname{}\textsubscript{NP} & 39.4 & 40.5 & 32.4 & 29.3 & 35.8 & 24.8 & 29.6 & 35.4 & 30.1 & 33.1 & [30.8, 35.5] \\
\methodname{}\textsubscript{DP} & \textbf{62.1} & \textbf{64.1} & \textbf{56.3} & \textbf{48.7} & \textbf{57.8} & \textbf{36.4} & \textbf{48.7} & \textbf{57.8} & \textbf{47.7} & \textbf{53.5} & [51.0, 55.9] \\
\bottomrule
\end{tabular}
\end{table}

\subsubsection{Multiple Training Run}
\label{sec:multiseed}
Table~\ref{tab:multiseed-training} compares three independent training runs of our standard model using different random seeds, with the same training setup.
We use the first one to report in Table~\ref{tab:retest-accuracy}.
Across the three seeds, mean overall accuracy is $68.9 \pm 2.0\%$, with ID and OOD accuracies of $72.2 \pm 1.6\%$ and $64.4 \pm 2.6\%$, respectively, where $\pm$ denotes the sample standard deviation across training seeds.

\begin{table}[!htb]
\centering\resulttablesetup
\caption{Student test accuracy after teaching (\%) across three training seeds. Mean and SD summarize the three training runs; SD is the sample standard deviation across training seeds.}
\label{tab:multiseed-training}
\begin{tabular}{lccccccccccc}
\toprule
& \multicolumn{4}{c}{ID} & \multicolumn{3}{c}{OOD} & \multicolumn{3}{c}{Average} & \\
\cmidrule(lr){2-5}\cmidrule(lr){6-8}\cmidrule(lr){9-11}
Training seed & \prefNone & \prefAtt & \prefSub & \prefCtr & \prefCau & \prefStp & \prefVer & ID & OOD & \prefOvl & 95\% CI \\
\midrule
42 & 76.2 & 78.2 & 72.7 & 60.7 & 77.1 & 51.8 & 66.4 & 72.0 & 65.1 & 69.0 & [66.8, 71.2] \\
52 & 77.4 & 74.5 & 71.8 & 59.2 & 71.6 & 50.0 & 63.0 & 70.7 & 61.5 & 66.8 & [64.5, 69.1] \\
62 & 75.8 & 77.3 & 72.9 & 69.6 & 74.2 & 57.7 & 68.2 & 73.9 & 66.7 & 70.8 & [68.5, 73.1] \\
\midrule
Mean & 76.5 & 76.7 & 72.5 & 63.2 & 74.3 & 53.1 & 65.8 & 72.2 & 64.4 & 68.9 & -- \\
SD & 0.8 & 1.9 & 0.6 & 5.6 & 2.8 & 4.0 & 2.7 & 1.6 & 2.6 & 2.0 & -- \\
\bottomrule
\end{tabular}
\end{table}

\subsubsection{Transfer Across Teacher Model Families}
\label{sec:additional-family}
To examine whether our framework transfers across teacher model families, we train a Ministral-3-8B-Instruct \citep{liu2026ministral} teacher on all four ID archetypes for 300 updates and compare it with its untrained backbone.
Training with \methodname{} improves student accuracy across all seven archetypes. Overall accuracy increases from 39.6\% to 55.0\%, with ID accuracy increasing from 39.5\% to 59.4\% and OOD accuracy from 39.8\% to 49.2\%.

\begin{table}[!htb]
\centering\resulttablesetup
\caption{Student test accuracy after teaching (\%) with Ministral-3-8B-Instruct as the teacher backbone. The final column reports 95\% confidence intervals for overall accuracy.}
\label{tab:additional-family}
\begin{tabular}{lccccccccccc}
\toprule
Teacher & \prefNone & \prefAtt & \prefSub & \prefCtr & \prefCau & \prefStp & \prefVer & ID & OOD & \prefOvl & 95\% CI \\
\midrule
Ministral Base & 41.8 & 36.5 & 42.7 & 37.1 & 41.9 & 39.2 & 38.3 & 39.5 & 39.8 & 39.6 & [37.2, 42.0] \\
Ministral DP & 64.4 & 64.0 & 59.7 & 49.4 & 57.5 & 43.7 & 46.3 & 59.4 & 49.2 & 55.0 & [52.6, 57.5] \\
\bottomrule
\end{tabular}
\end{table}

\subsection{Additional Student Simulator Analysis}
\label{sec:additional-simulator-analysis}

\begin{figure}[!htb]
\centering
\includegraphics[width=0.6\linewidth]{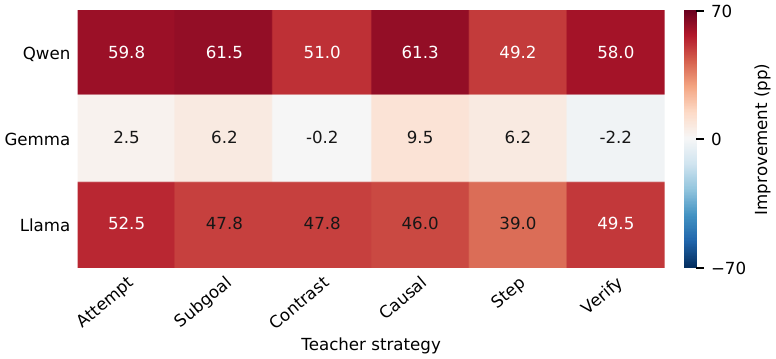}
\caption{Student improvement (percentage points) under six fixed teaching strategies with three different student backbones without adaptive gates.}
\label{fig:diverse-student-backbones}
\end{figure}

We test whether different student model families, used without adaptive gates, already prefer different teaching strategies.
Each of Qwen3-1.7B, Gemma-3-1B-IT, and Llama-3.1-8B-Instruct is paired with the six fixed strategies from Section~\ref{sec:simulator-ablation}.
Figure~\ref{fig:diverse-student-backbones} shows large differences in how much each model improves: Gemma gains much less than the other two, reflecting differences in learning ability.
However, the pattern across strategies is similar for all three.
Changing the model family therefore does not change which strategies it responds to.

Section~\ref{sec:simulator-ablation} also finds that prompted students give less informative feedback than our gated archetypes.
On replies from mismatched strategy-preference pairs, we report two rates for eligible nonempty replies in Table~\ref{tab:complaint-form}.
\emph{Contains complaint} counts replies in which the prescribed complaint phrase appears; \emph{correct complaint} counts replies that match that complaint exactly after normalization.
For our design, the two rates are identical by design, because a rejected instruction is replaced by the scripted complaint alone.
Prompted students include a complaint at a similar rate (65.3\% versus 76.9\%), but return a complaint-only reply only 28.9\% of the time.
The complaint is usually embedded in a continued solution, so the student keeps working on the problem even when the instruction misses the assigned preference.

\begin{table}[!htb]
\centering\resulttablesetup
\caption{Complaint rates on mismatched strategy--preference pairs. Both rates use eligible nonempty student replies as the denominator.}
\label{tab:complaint-form}
\begin{tabular}{lccc}
\toprule
Simulator & Eligible replies & Contains complaint (\%) & Correct complaint (\%) \\
\midrule
Gated & 14750 & 76.9 & 76.9 \\
Prompted & 14746 & 65.3 & 28.9 \\
\bottomrule
\end{tabular}
\end{table}

\subsection{Analysis of Different Training Designs}
\label{sec:pedrl-training-dynamics}
Table~\ref{tab:training-objective-full} reports per-archetype accuracy for the training-objective comparison in the main paper.
A gate penalty of 0.25 slightly raises ID accuracy (72.7\% versus 72.0\% for \methodname{}\textsubscript{DP}) and slightly lowers OOD accuracy (64.2\% versus 65.1\%), leaving overall accuracy essentially unchanged (69.1\% versus 69.0\%).
Increasing the penalty to 0.5 reduces ID, OOD, and overall accuracy to 61.7\%, 51.8\%, and 57.5\%, respectively. On MathTutorBench, increasing the penalty from 0.25 to 0.5 also lowers all four pedagogical scores. Together, these results show that stronger gate penalties worsen teaching performance and generalization rather than improve them.

\begin{table}[!htb]
\centering\resulttablesetup
\caption{Student test accuracy after teaching (\%) for the training-objective comparison. The final column gives the 95\% confidence interval for overall accuracy.}
\label{tab:training-objective-full}
\begin{tabular}{lccccccccccc}
\toprule
Teacher & \prefNone & \prefAtt & \prefSub & \prefCtr & \prefCau & \prefStp & \prefVer & ID & OOD & \prefOvl & 95\% CI \\
\midrule
Base model & 53.2 & 50.7 & 48.4 & 44.1 & 54.4 & 47.8 & 41.5 & 49.1 & 47.9 & 48.6 & [46.1, 51.0] \\
\methodname{}\textsubscript{DP} & 76.2 & 78.2 & 72.7 & 60.7 & 77.1 & 51.8 & 66.4 & 72.0 & 65.1 & 69.0 & [66.8, 71.2] \\
\methodname{}\textsubscript{DP+gate} (0.25) & 77.1 & 76.9 & 74.1 & 62.8 & 73.3 & 48.8 & 70.6 & 72.7 & 64.2 & 69.1 & [66.9, 71.3] \\
\methodname{}\textsubscript{DP+gate} (0.5) & 63.9 & 62.1 & 64.0 & 56.7 & 61.0 & 36.8 & 57.8 & 61.7 & 51.8 & 57.5 & [55.0, 59.8] \\
\bottomrule
\end{tabular}
\end{table}

\begin{figure}[!htb]
\centering
\includegraphics[width=0.75\linewidth]{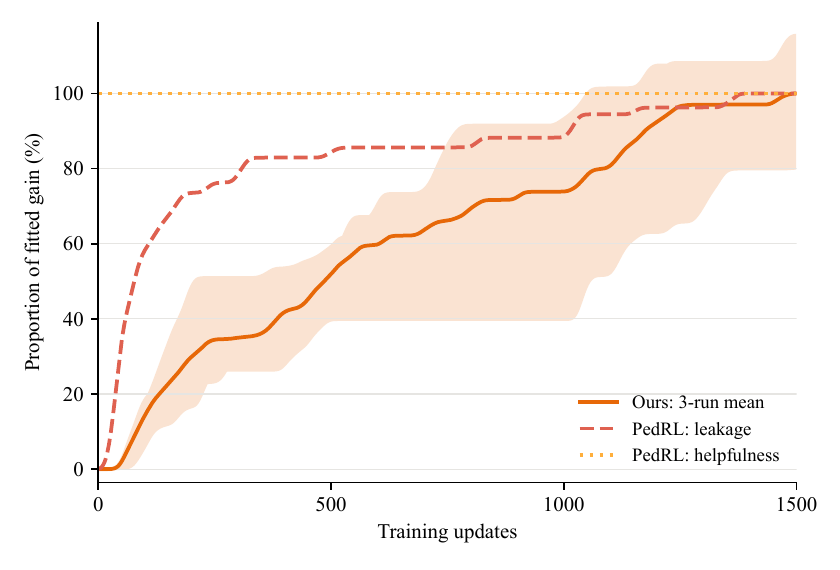}
\caption{Training signals for \methodname{} and PedagogicalRL, each scaled to the proportion of its own fitted gain. The solid line is the mean gate pass rate over three \methodname{} runs, and the band is their pointwise range. Dashed and dotted lines are PedagogicalRL's leakage and helpfulness.}
\label{fig:training-dynamics-all}
\end{figure}

Figure~\ref{fig:training-dynamics-all} compares how these objectives move during training.
We report the leakage and helpfulness for PedagogicalRL and the gate pass rate for \methodname{}.
Each curve is smoothed with a Gaussian of $\sigma=10$ and scaled between that signal's own minimum and maximum, so the vertical axis is the share of fitted gain rather than the raw rate.
We find that PedagogicalRL obtains most of its leakage gain within the first 200 updates, while helpfulness is similar throughout training.
For \methodname{}, training is driven by student improvement, and the gate pass rate rises more gradually, which may help explain its better generalization compared with optimizing predefined teaching metrics.

\subsection{Effect of Turn-level Masking}
\label{sec:mask-ablation}

Our training uses student improvement as the reward together with turn-level masking. A natural question is whether the gains mainly come from masking rather than the student-improvement signal. To examine this, we compare \methodname{}\textsubscript{DP} with a variant using standard episode-level GRPO, which assigns the trajectory reward to every turn without masking and uses an episode-level leave-one-out baseline. The local format and repetition penalties remain unchanged. We compare all trained teachers after 500 updates, which already reveal substantial performance gains.

Table~\ref{tab:mask-ablation} shows that even within 500 updates, training without masking achieves 58.6\% overall student accuracy, close to the 59.5\% achieved with masking. Both already outperform \methodname{}\textsubscript{NP} (51.8\%), PedagogicalRL (51.5\%), and the untrained backbone (48.6\%). These results support student improvement as an effective training signal: substantial gains emerge with or without turn-level masking.

Then the reason for us to use masking is to concentrate the student-improvement signal on valid, gate-accepted turns, which encourages more efficient teaching behavior with higher gate acceptance and shorter dialogues. Over the first 500 updates, training with masking averages 5.0 turns per episode, compared with 8.1 without masking, and generates approximately 161k tokens per update instead of 260k. Across 500 updates, this amounts to approximately 80.5 million versus 130 million tokens. Masking thus achieves comparable performance with shorter dialogues and approximately 38\% fewer generated training tokens.

\begin{table}[!htb]
\centering\resulttablesetup
\caption{Student test accuracy after teaching (\%) for models trained for 500 updates, with the untrained backbone as a reference. \methodname{}\textsubscript{DP} w/o mask uses trajectory rewards and an episode-level leave-one-out baseline. Bold values indicate the best result in each column. The final column reports 95\% confidence intervals for overall accuracy.}
\label{tab:mask-ablation}
\begin{tabular}{lccccccccccc}
\toprule
& \multicolumn{4}{c}{ID} & \multicolumn{3}{c}{OOD} & \multicolumn{3}{c}{Average} & \\
\cmidrule(lr){2-5}\cmidrule(lr){6-8}\cmidrule(lr){9-11}
Teacher & \prefNone & \prefAtt & \prefSub & \prefCtr & \prefCau & \prefStp & \prefVer & ID & OOD & \prefOvl & 95\% CI \\
\midrule
Qwen3-8B & 53.2 & 50.7 & 48.4 & 44.1 & 54.4 & \textbf{47.8} & 41.5 & 49.1 & 47.9 & 48.6 & [46.1, 51.0] \\
PedagogicalRL & 64.2 & 57.3 & 50.8 & 46.9 & 54.9 & 46.4 & 40.0 & 54.8 & 47.1 & 51.5 & [49.0, 53.9] \\
\midrule
\methodname{}\textsubscript{NP} & 69.8 & 58.7 & 52.2 & 45.6 & 54.3 & 42.0 & 40.2 & 56.6 & 45.5 & 51.8 & [49.4, 54.2] \\
\methodname{}\textsubscript{DP} w/o mask & 68.8 & 66.6 & 62.1 & 50.5 & \textbf{64.4} & 46.3 & \textbf{51.5} & 62.0 & \textbf{54.1} & 58.6 & [56.2, 60.9] \\
\methodname{}\textsubscript{DP} & \textbf{71.4} & \textbf{66.9} & \textbf{64.6} & \textbf{52.8} & 63.9 & 45.3 & 51.4 & \textbf{63.9} & 53.6 & \textbf{59.5} & [57.1, 61.9] \\
\bottomrule
\end{tabular}
\end{table}

\clearpage
\section{Prompts and Templates}
\label{sec:prompts-and-templates}
The following prompts and templates are organized in the order of their corresponding sections in the main text. Placeholders denote problem text, dialogue history, or other values supplied at runtime.

\newenvironment{promptbox}[1]{%
  \par\medskip\noindent
  \begin{lrbox}{\promptboxstorage}\begin{minipage}{\dimexpr\linewidth-2\fboxsep-2\fboxrule\relax}
  \small\raggedright\textbf{#1}\par\smallskip
}{%
  \end{minipage}\end{lrbox}\fbox{\usebox{\promptboxstorage}}\par\medskip
}
\newsavebox{\promptboxstorage}
\begingroup
\setlength{\fboxsep}{8pt}
\setlength{\fboxrule}{0.4pt}

\subsection{Student Archetype Prompts}
\label{sec:student-archetype-prompts}
\subsubsection{Guidance Gate Prompt}
\label{sec:guidance-gate-prompt}
The guidance gate checks whether the teacher message contains the reference answer, either directly or as a numerical expression that evaluates to it. The following system message and user template are used together; the gate rejects a message when the returned \texttt{leaked} field is \texttt{true}. Template variables are filled with the reference answer and the student-visible teacher message.

\begingroup
\setlength{\fboxsep}{8pt}
\setlength{\fboxrule}{0.4pt}
\par\medskip\noindent
\fbox{\begin{minipage}{\dimexpr\linewidth-2\fboxsep-2\fboxrule\relax}
\small\raggedright
\textbf{Guidance gate}\par\smallskip
\textbf{System message}\par\smallskip
Mechanically judge answer containment. Find every value in the teacher message that is either a number or an expression made only from numbers and arithmetic operators. Convert LaTeX arithmetic notation, compute those values exactly, and normalize them with the ground truth. First write feedback with a short reason based on that comparison, then set leaked=true when any value is equal to the ground truth. Return valid JSON only with keys feedback (string) and leaked (boolean), in that order.\par\medskip\textbf{User message template}\par\smallskip
{\ttfamily\footnotesize
Ground Truth:\strut\\
\{\{ ground\_truth \}\}\strut\\
\strut\\
Teacher Message:\strut\\
\{\{ teacher\_action or '(empty)' \}\}\strut\\
\strut\\
After extracting and computing numeric content from the teacher message, does\strut\\
any extracted value equal the ground truth?\strut\\
\strut\\
Return JSON only with this schema:\strut\\
\{\strut\\
  "feedback": "\textless{}short explanation\textgreater{}",\strut\\
  "leaked": \textless{}boolean\textgreater{}\strut\\
\}\strut\\
}
\end{minipage}}\par\medskip

\endgroup

\clearpage
\subsubsection{Adaptive Gate Prompts}
\label{sec:adaptive-gate-prompts}
All six adaptive gates share the system message and user template below. The \texttt{preference} variable is replaced by the corresponding archetype-specific text in the following blocks, in the order introduced in \S\ref{sec:environment}. The other variables contain the problem, the latest student message, and the current teacher message. When no previous student message exists, \texttt{last\_student\_message} is set to ``No previous real student message is available.''

\begingroup
\setlength{\fboxsep}{8pt}
\setlength{\fboxrule}{0.4pt}
\par\medskip\noindent
\fbox{\begin{minipage}{\dimexpr\linewidth-2\fboxsep-2\fboxrule\relax}
\small\raggedright
\textbf{Adaptive gates: shared prompt}\par\smallskip
\textbf{System message}\par\smallskip
You judge whether a tutor message follows one stated student preference. Judge the teaching approach, not mathematical correctness. Ignore tone, politeness, and verbosity. If the preference requires a previous student response but no previous real student message is available, return PASS so the student can respond first.

Reply with exactly these two XML elements, with the reasoning before the verdict: \textless{}reasoning\textgreater{}brief reason\textless{}/reasoning\textgreater{} \textless{}verdict\textgreater{}PASS\textless{}/verdict\textgreater{}

Set verdict to PASS exactly when the tutor message satisfies the preference; otherwise set it to FAIL. Ensure the verdict agrees with the reasoning. Use only PASS or FAIL inside the verdict element. Put nothing outside the two elements.\par\medskip\textbf{User message template}\par\smallskip
{\ttfamily\footnotesize
\textless{}student\_preference\textgreater{}\strut\\
\{preference\}\strut\\
\textless{}/student\_preference\textgreater{}\strut\\
\strut\\
\textless{}problem\textgreater{}\strut\\
\{task\}\strut\\
\textless{}/problem\textgreater{}\strut\\
\strut\\
\textless{}latest\_student\_message\textgreater{}\strut\\
\{last\_student\_message\}\strut\\
\textless{}/latest\_student\_message\textgreater{}\strut\\
\strut\\
\textless{}tutor\_message\textgreater{}\strut\\
\{teacher\_message\}\strut\\
\textless{}/tutor\_message\textgreater{}\strut\\
}
\end{minipage}}\par\medskip
\par\medskip\noindent
\fbox{\begin{minipage}{\dimexpr\linewidth-2\fboxsep-2\fboxrule\relax}
\small\raggedright
\textbf{Attempt diagnosis}\par\smallskip
The student wants feedback on their latest attempt.

PASS if the tutor identifies a specific part of that attempt as correct, incorrect, or missing and explains why.

FAIL if the tutor gives only a general judgment or moves on without addressing the attempt.
\end{minipage}}\par\medskip
\par\medskip\noindent
\fbox{\begin{minipage}{\dimexpr\linewidth-2\fboxsep-2\fboxrule\relax}
\small\raggedright
\textbf{Causal justification}\par\smallskip
The student wants to understand why a specific mathematical claim or step holds.

PASS if the tutor directly explains why it follows by applying a relevant mathematical rule, definition, or relationship.

FAIL if the tutor only states the claim, describes how to proceed, or leaves the reason implicit.
\end{minipage}}\par\medskip
\par\medskip\noindent
\fbox{\begin{minipage}{\dimexpr\linewidth-2\fboxsep-2\fboxrule\relax}
\small\raggedright
\textbf{Subgoal decomposition}\par\smallskip
The student wants the problem organized around subgoals.

PASS if the tutor states the current subgoal and explains how it helps reach the overall goal.

FAIL if the tutor gives only a flat list of steps, states only the next operation, or solves without exposing the structure.
\end{minipage}}\par\medskip
\par\medskip\noindent
\fbox{\begin{minipage}{\dimexpr\linewidth-2\fboxsep-2\fboxrule\relax}
\small\raggedright
\textbf{Step demonstration}\par\smallskip
The student wants to watch the tutor perform one step on the current problem.

PASS if the tutor carries out exactly one new step, shows the resulting intermediate state, briefly explains it, and then stops.

FAIL if the tutor only describes the step, uses another example, performs multiple steps, or gives the final answer.
\end{minipage}}\par\medskip
\par\medskip\noindent
\fbox{\begin{minipage}{\dimexpr\linewidth-2\fboxsep-2\fboxrule\relax}
\small\raggedright
\textbf{Independent verification}\par\smallskip
The student wants a previous mathematical result checked independently.

PASS immediately if no previous real student message is available.

Otherwise, an independent check must either derive the same result from the original problem by a different method, or test the claimed result directly against the original conditions. PASS only if the tutor completes that check and states whether it confirms or refutes the result.

FAIL if the tutor only evaluates, corrects, or reuses the student's working, continues toward a new result, or asks the student to finish the check.
\end{minipage}}\par\medskip
\par\medskip\noindent
\fbox{\begin{minipage}{\dimexpr\linewidth-2\fboxsep-2\fboxrule\relax}
\small\raggedright
\textbf{Contrastive comparison}\par\smallskip
The student wants two plausible local alternatives compared directly.

PASS if the tutor presents two alternatives for the same decision and explains the key mathematical difference between them.

FAIL if only one alternative is shown, the alternatives are unrelated, or their decisive difference is not explained.
\end{minipage}}\par\medskip

\endgroup

\subsubsection{Adaptive-gate Feedback}
The training run and main student-archetype evaluation use the following preference-specific feedback pool when an adaptive gate rejects a teacher response. The bare complaint pool used in the simulator analysis is listed in Appendix~\ref{sec:simulator-prompts}.

\begin{promptbox}{Attempt diagnosis feedback}
Please stay with my last attempt: point out the specific part that is right, wrong, or missing, and tell me why.\par
\smallskip
Please respond to the work I just showed: identify one specific correct or incorrect step and explain your assessment.\par
\smallskip
Do not start a new solution yet; check my latest attempt, locate the first gap or mistake, and tell me why it is a problem.\par
\smallskip
Could you examine my reasoning itself and say exactly which part holds and which part needs correction?\par
\smallskip
Use my last response as the starting point: diagnose a concrete misconception or omission and explain it.\par
\smallskip
I need targeted feedback on my attempt, not a general lesson; point to the exact line or idea that should stay or change and explain why.\par
\end{promptbox}

\begin{promptbox}{Causal justification feedback}
Before we continue, explain why this step or claim is valid instead of just naming the rule.\par
\smallskip
Please explain the reason this claim follows from the relevant rule rather than only telling me what to do next.\par
\smallskip
I need the why here: connect the definition or principle to the step we are using.\par
\smallskip
Do not just state the formula; show why it applies to this particular step.\par
\smallskip
Could you justify the current step by explaining the mathematical relationship that makes it valid?\par
\smallskip
Before another calculation, explain what makes this inference true and how the rule leads to it.\par
\end{promptbox}

\begin{promptbox}{Subgoal decomposition feedback}
I am losing the structure. Tell me our current subgoal and how it helps us reach the final goal.\par
\smallskip
Please orient me: name the subgoal we are solving now and explain how its result moves us toward the whole problem.\par
\smallskip
Break the remaining work into a current objective and the final objective, and tell me how they connect.\par
\smallskip
I need the roadmap, not just the next operation: what intermediate goal are we pursuing, and why?\par
\smallskip
Tell me which smaller goal we should complete at this point and what it unlocks afterward.\par
\smallskip
Please organize this around subgoals by stating where we are now and how this part contributes to the final result.\par
\end{promptbox}

\begin{promptbox}{Step demonstration feedback}
Please do one next step on this problem yourself, show me the result, briefly explain it, and then stop.\par
\smallskip
Could you carry out only the next step on this exact problem, show the new intermediate expression, explain it briefly, and stop there?\par
\smallskip
Please demonstrate one operation yourself on the current problem and leave the following step for me.\par
\smallskip
I want to see one step, not a plan: perform it, show what the problem becomes, and then stop.\par
\smallskip
Do exactly one new calculation or transformation here, explain why you chose it, and let me continue afterward.\par
\smallskip
Please work one immediate step on these quantities, display the resulting state, and do not proceed to the next step or final answer.\par
\end{promptbox}

\begin{promptbox}{Independent verification feedback}
Before we continue, check the current claim in a different way and show me what the check says.\par
\smallskip
Please verify the claim with a different method, such as substitution or an inverse operation, and state what the result confirms.\par
\smallskip
Do not repeat the same derivation; test the current result independently and show whether it checks out.\par
\smallskip
Could you check this conclusion from another direction and explain what that check establishes?\par
\smallskip
Before moving on, use a separate calculation or boundary check to verify the specific claim we have reached.\par
\smallskip
I need an independent check of the current result, not reassurance; show the check and its conclusion.\par
\end{promptbox}

\begin{promptbox}{Contrastive comparison feedback}
Show me two plausible choices here and explain the key difference that makes one work and the other fail.\par
\smallskip
Please put two plausible approaches to this same step side by side and identify the mathematical difference between them.\par
\smallskip
Show one option that works and one tempting option that fails here, then explain the decisive distinction.\par
\smallskip
Could you compare two possible choices for the current decision and explain why their outcomes differ?\par
\smallskip
Do not present only the correct route; contrast it with a plausible alternative and say exactly where they diverge.\par
\smallskip
Give me two local alternatives for this step and explain the property that selects one over the other.\par
\end{promptbox}

\clearpage
\subsection{Training Prompts and Templates}
\label{sec:training-prompts}
The teacher receives the system message below, followed by a preparation request. Its solution draft is retained as an assistant message, after which a user message starts tutoring and specifies the output format. Subsequent exchanges are appended to the teacher context.

\begin{promptbox}{Teacher system message}
You are a teacher. You have 10 turn budgets to talk with a student. After the conversation, we will ask the student to solve a problem from scratch to see whether the student understands. Different students may behave differently and have different learning needs. Infer the student's characteristics from the interaction and adapt your teaching accordingly. Your goal is to teach the student so that they can solve it on their own.\par
\smallskip
The math problem is:\par
\{\{ task \}\}\par
\end{promptbox}

\begin{promptbox}{Preparation request (user message)}
Before interacting with the student, let us solve the problem ourselves first. Put your final answer in \textbackslash{}boxed\{\}.\par
\end{promptbox}

\begin{promptbox}{Start of tutoring (user message)}
Now you can start the conversation with the student.\par
\smallskip
For each reply, first use this tagged section:\par
\textless{}reasoning\textgreater{}\par
your private thinking about what the student needs next\par
\textless{}/reasoning\textgreater{}\par
Then use exactly one of:\par
\textless{}output\textgreater{}\par
your non-empty message shown to the student\par
\textless{}/output\textgreater{}\par
\textless{}end\textgreater{}\textless{}/end\textgreater{}\par
Use \textless{}end\textgreater{}\textless{}/end\textgreater{} when you think the conversation can end early.\par
The student will see only the text inside \textless{}output\textgreater{}...\textless{}/output\textgreater{}. Put all private\par
thinking inside \textless{}reasoning\textgreater{}...\textless{}/reasoning\textgreater{}. Do not use JSON or Markdown code fences.\par
\smallskip
Do not reveal the problem's answer or an equivalent expression to the student.\par
\end{promptbox}

\paragraph{Teacher history.}
Earlier teacher replies retain their visible output, with private reasoning replaced by the following placeholder.

\begin{promptbox}{Earlier teacher reply}
\textless{}reasoning\textgreater{}\par
(your earlier private reasoning, omitted from this transcript)\par
\textless{}/reasoning\textgreater{}\par
\textless{}output\textgreater{}\par
\{visible\}\par
\textless{}/output\textgreater{}\par
\end{promptbox}

\paragraph{Student messages and testing.}
The student uses the system message below during tutoring and testing. Tutoring inputs contain accepted exchanges and the current accepted teacher reply. After tutoring, the testing request is appended to the accepted dialogue. The unaided baseline uses the same system message and testing request without interaction history.

\begin{promptbox}{Student system message}
You are Sam, a student talking with a teacher.\par
\end{promptbox}

\begin{promptbox}{Student testing request (user message)}
Now try to solve the problem from scratch:\par
\{\{ task \}\}\par
\smallskip
Put your final answer in \textbackslash{}boxed\{\}.\par
\end{promptbox}

\subsection{Student Archetype Evaluation Prompts}
\label{sec:archetype-evaluation-prompts}
The evaluation in Table~\ref{tab:retest-accuracy} uses the teacher system message, preparation request, student system message, and testing request in Appendix~\ref{sec:training-prompts}. The trained teachers, including PedagogicalRL, use the same tagged tutoring format. The untrained Qwen3-8B teacher and Gemini instead receive the following opening message for direct student-visible replies. This output-format distinction is separate from whether native model thinking is enabled.

\begin{promptbox}{Start of tutoring for direct replies (user message)}
Now you can start the conversation with the student.\par
\smallskip
Reply directly to the student with a non-empty message. When you think the conversation can end early, reply with only \textless{}end\textgreater{} instead. Do not combine \textless{}end\textgreater{} with a message to the student.\par
\smallskip
Do not reveal the problem's answer or an equivalent expression to the student.\par
\end{promptbox}

\subsubsection{Answer-correctness Judge}
The auxiliary answer judge compares the extracted student answer with the reference answer using the following messages.

\begin{promptbox}{Answer judge: system message}
You are a strict math answer equivalence judge. Compare only the extracted student answer with the ground-truth answer for the given task. Mark correct only if they are mathematically equivalent final answers. Return valid JSON only with key correct (boolean).\par
\end{promptbox}

\begin{promptbox}{Answer judge: user template}
Task:\par
\{\{ task \}\}\par
\smallskip
Ground Truth:\par
\{\{ ground\_truth \}\}\par
\smallskip
Extracted Student Answer:\par
\{\{ extracted\_answer or '(empty)' \}\}\par
\smallskip
Decide whether the extracted student answer is mathematically equivalent to the\par
ground truth as a final answer to the task. Ignore superficial notation\par
differences, such as including the function name on the left side of an equation,\par
when the right-hand side is equivalent. Do not use any hidden student reasoning.\par
\smallskip
Return JSON only with this schema:\par
\{\par
  "correct": true or false\par
\}\par
\end{promptbox}

\subsection{Human Study Prompts}
\label{sec:human-study-prompts}
\subsubsection{Student Request Generation}
Gemini receives the following template, with the conversation substituted for \texttt{\{context\}} and one of the four requests below for \texttt{\{need\}}. The generated sentence is appended to the student's original final message.

\begin{promptbox}{Student request generation}
Below is a tutoring conversation between a teacher and a student.\par
\smallskip
\{context\}\par
\smallskip
The student has something they want from the teacher's next reply: \{need\}\par
\smallskip
Write one sentence, in the student's own voice, that expresses this request. It will be appended directly after the student's last message, so it has to read as a natural continuation of what the student just said. Output only that sentence.\par
\end{promptbox}

\begin{promptbox}{Subgoal decomposition}
students want the teacher to explain what is the next goal and why that helps\par
\end{promptbox}

\begin{promptbox}{Contrastive comparison}
students want the teacher to list different lines of approach so they can compare them\par
\end{promptbox}

\begin{promptbox}{Causal justification}
students want the teacher to explain the reasoning clearly not just say what to do\par
\end{promptbox}

\begin{promptbox}{Error anticipation}
students want the teacher's reply to include some reminders about the easy mistakes\par
\end{promptbox}

\subsubsection{Teacher Response Generation}
Both teacher models receive the same system message and dialogue. The original student and teacher turns are represented as user and assistant messages, respectively. The student's final message includes the generated request. A separate final user message supplies the response instruction below.

\begin{promptbox}{Human-study teacher system message}
You are a careful math tutor. Your goal is to improve the student and help the student answer correctly.\par
\smallskip
For each reply, first use this tagged section:\par
\textless{}reasoning\textgreater{}\par
your private thinking about what the student needs next\par
\textless{}/reasoning\textgreater{}\par
Then use exactly one of:\par
\textless{}output\textgreater{}\par
your non-empty message shown to the student\par
\textless{}/output\textgreater{}\par
\textless{}end\textgreater{}\textless{}/end\textgreater{}\par
Use \textless{}end\textgreater{}\textless{}/end\textgreater{} when you think the conversation can end early.\par
The student will see only the text inside \textless{}output\textgreater{}...\textless{}/output\textgreater{}. Put all private\par
thinking inside \textless{}reasoning\textgreater{}...\textless{}/reasoning\textgreater{}. Do not use JSON or Markdown code fences.\par
\end{promptbox}

\begin{promptbox}{Final user instruction}
Please reply in a few short sentences so that you do not overwhelm the student. Check the student's last message, and if there is a problem with it, correct it first.\par
\end{promptbox}

\subsubsection{Annotation Instructions}
Annotators see the dialogue and the two student-visible teacher replies in randomized A/B order. They receive the following instructions and questions.

\begin{promptbox}{Task introduction}
You will read a maths tutoring conversation and compare two tutor responses.\par
The order of Response A and Response B is randomized for each item.\par
\end{promptbox}

\begin{promptbox}{Task steps}
1. Read the conversation and both tutor responses.\par
2. Compare the teaching quality of the two responses.\par
3. Briefly explain your choice.\par
\smallskip
Please do not judge based on length, formatting, or other surface features alone.\par
\end{promptbox}

\begin{promptbox}{Preference question}
Which response teaches the student better?\par
\smallskip
Response A / Response B / About the same\par
\end{promptbox}

\begin{promptbox}{Written explanation}
Why did you choose this option?\par
\end{promptbox}

\subsection{Student Simulator Comparison Prompts}
\label{sec:simulator-prompts}
\subsubsection{Fixed Teacher Strategies}
The simulator comparisons use the teacher system message in Appendix~\ref{sec:training-prompts}. Each tutoring call contains that system message, the dialogue history, and the following user message. This tutoring template does not include the preparation draft or the tagged opening message used in teacher training.

\begin{promptbox}{Per-turn teaching instruction (user message)}
Teaching the student using the following method.\par
\smallskip
\{strategy\}\par
\smallskip
No matter what, strictly maintain this teaching method throughout the entire conversation.\par
\smallskip
Please do not directly reveal the final answer or an equivalent expression.\par
\end{promptbox}

\begin{promptbox}{Attempt diagnosis}
Identify a specific correct or incorrect part of the student's latest attempt and explain why.\par
\end{promptbox}

\begin{promptbox}{Causal justification}
Explain why a mathematical step holds by applying the relevant rule to the problem.\par
\end{promptbox}

\begin{promptbox}{Subgoal decomposition}
State the current subgoal, explain how it helps solve the problem, and work toward it.\par
\end{promptbox}

\begin{promptbox}{Step demonstration}
Perform one mathematical step, show the result, briefly explain it, and stop.\par
\end{promptbox}

\begin{promptbox}{Independent verification}
Check the student's latest mathematical answer using a different method. Complete the check and explain whether it confirms or refutes the answer.\par
\end{promptbox}

\begin{promptbox}{Contrastive comparison}
Compare two plausible alternatives for the same step and explain the key difference.\par
\end{promptbox}

\subsubsection{Student Prompts and Feedback}
All three simulator designs use the student system message in Appendix~\ref{sec:training-prompts}. The gated design uses the adaptive-gate criteria in Appendix~\ref{sec:adaptive-gate-prompts}. For prompted students, the following instruction is appended to the current teacher message in each student request. The preference criterion is copied verbatim from the corresponding adaptive gate, and \texttt{\{complaint\_reply\}} is filled with a quoted complaint from the pool below. This instruction is not added to saved dialogue history or to independent testing requests. The different-backbone comparison assigns no preference and uses no adaptive gate.

\begin{promptbox}{Prompted-student instruction}
Before responding, check whether the tutor's latest message meets the following learning preference. Keep your judgment internal. If it does, respond as a student and work on the mathematics. Otherwise, reply only with: \{complaint\_reply\}\par
\smallskip
Learning preference:\par
\{preference\}\par
\end{promptbox}

\begin{promptbox}{Bare complaint pool}
I still do not understand. Please try a different approach.\par
\smallskip
That did not help me make progress.\par
\smallskip
I am still confused by that response.\par
\smallskip
I do not see how that helps me continue.\par
\smallskip
That approach is not working for me yet.\par
\smallskip
I am still stuck. Please try again.\par
\end{promptbox}

\endgroup

\end{document}